\documentclass[sigconf]{acmart}

\AtBeginDocument{%
  }

\copyrightyear{2026}
\acmYear{2026}
\setcopyright{cc}
\setcctype{by}
\acmConference[MM '26]{Proceedings of the 34th ACM International Conference on Multimedia}{November 10--14, 2026}{Rio de Janeiro, Brazil}
\acmBooktitle{Proceedings of the 34th ACM International Conference on Multimedia (MM '26), November 10--14, 2026, Rio de Janeiro, Brazil}
\acmDOI{10.1145/3767308.3836313}
\acmISBN{979-8-4007-2213-4/2026/11}
\usepackage{microtype}
\usepackage{cleveref}
\usepackage{amsmath}
\usepackage[most]{tcolorbox}
\usepackage{tabularx}

\begin{document}
\settopmatter{printfolios=true}
\title{Towards Expressive and Faithful Audio-to-Image Generation: A Unified Multimodal Dataset and Synthesis Framework}

\author{Dongxu Ge}
\orcid{0009-0009-0560-8270}
\authornote{Work done during an internship at TeleAI.}
\affiliation{%
  \institution{University of Science and Technology of China}
  \city{Hefei}
  \country{China}
}
\affiliation{%
  \institution{Institute of Artificial Intelligence, China Telecom (TeleAI)}
  \city{Beijing}
  \country{China}
}
\email{gedongxu@mail.ustc.edu.cn}

\author{Shansong Liu}
\orcid{0000-0001-6202-5615}
\affiliation{%
  \institution{Institute of Artificial Intelligence, China Telecom (TeleAI)}
  \city{Beijing}
  \country{China}
}
\email{dadinghh2@gmail.com}

\author{Cheng Gong}
\orcid{0009-0004-0272-3541}
\affiliation{%
  \institution{Institute of Artificial Intelligence, China Telecom (TeleAI)}
  \city{Beijing}
  \country{China}
}
\email{gongchengcheng@tju.edu.cn}

\author{Xiao-Lei Zhang}
\orcid{0000-0001-7694-193X}
\authornote{Corresponding authors.}
\affiliation{%
  \institution{Institute of Artificial Intelligence, China Telecom (TeleAI)}
  \city{Beijing}
  \country{China}
}
\affiliation{%
  \institution{Northwest Polytechnical University}
  \city{Xi'an}
  \country{China}
}

\email{xiaolei.zhang@nwpu.edu.cn}

\author{Chi Zhang}
\orcid{0009-0002-3514-2490}
\affiliation{%
  \institution{Institute of Artificial Intelligence, China Telecom (TeleAI)}
  \city{Beijing}
  \country{China}
}
\email{zhangc120@chinatelecom.cn}

\author{Xuelong Li}
\orcid{0000-0002-0019-4197}
\authornotemark[2]
\affiliation{%
  \institution{Institute of Artificial Intelligence, China Telecom (TeleAI)}
  \city{Beijing}
  \country{China}
}
\email{xuelong_li@ieee.org}

\renewcommand{\shortauthors}{Dongxu Ge. et al.}

\begin{abstract}
  As an important subfield of cross-modal generation, synthesizing static visual content in the form of images from audio, namely audio-to-image (A2I) generation, has attracted increasing research attention in recent years. Nevertheless, despite the remarkable visual quality of modern text-to-image (T2I) models, the performance of A2I remains fundamentally limited by traditional datasets, which often lack both high-fidelity images and precise cross-modal alignment. As a result, existing methods still struggle to achieve high-quality audio-to-image generation through finetuning strong T2I models, thereby constraining practical applications in this area. Motivated by this gap, we introduce A2I-Set, a unified, high-quality tri-modal dataset consisting of 323K paired audio, images, and detailed text captions, specifically designed for audio-visual research, including audio-conditioned image generation. Besides, we developed a new mixed-source test set for the A2I task through human supervision. We further propose an A2I model, AudioCanvas, fine-tuned on our A2I-Set. Experiments show that AudioCanvas achieves more visually expressive as well as cross-modal alignment results that generally outperforming existing approaches. Our dataset and source code are available at \url{https://github.com/gdx012/A2I-Generation}.
\end{abstract}




\maketitle

\begin{figure*}[!htbp]  
    \centering  
    \includegraphics[width=0.95\textwidth]{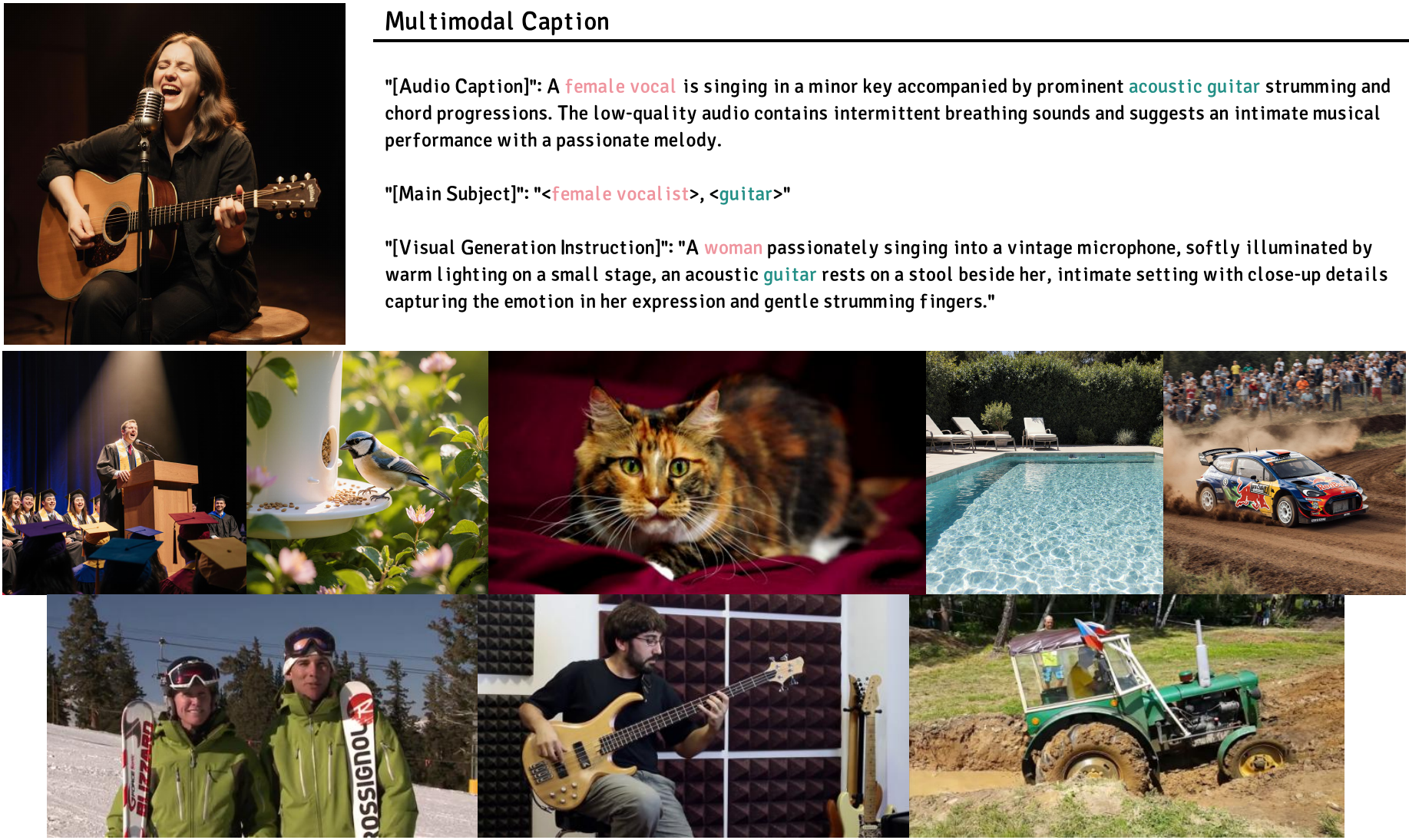}
    \caption{Some examples from A2I-Set. A2I-Set has a detailed description of audio as well as paired visual generation instructions Benefit from strict filter strategy and strong T2I model, our images are superior in terms of visual effects and audio-vision alignment.}  
    \label{fig:example}  
\end{figure*}

\section{Introduction}
Cross-modal generation, including the mutual generation of modalities such as vision, text, and audio, is a current research hotspot \cite{ruan2023mm, girdhar2023imagebind, zhao2025foleyspace, zhu2025viewmask, zhao2025uniform, shao2025ai, cheng2026unisonharmonizingmotionspeech, song2026interactiveavatarrealtimestreamingvideo}. As the most common form of visual expression, images are also the focus of cross-modal research. Extensive research has significantly accelerated text-to-image generation \cite{huang2026nfig}, and the ability of models to generate high-fidelity and diverse images has been significantly enhanced. Thanks to their powerful performance, text-to-image models have been widely applied in multiple commercial fields. Moreover, recently, many studies have begun to focus on audio-to-image generation. Although significant progress has been made, the expressiveness of audio-to-image generation is still difficult to compare with SOTA-level text-to-image models. We attribute this phenomenon to the following reasons: 

Firstly, due to relatively few studies on visual-audio modality earlier, there are few datasets focusing on visual-audio bimodality. Traditional datasets either only focus on the visual part and lack various classes of audio data, or pay more attention to audio quality and annotation accuracy while ignoring visual quality. The lack of datasets in the visual-audio field has hindered further research on the mutual generation of two modalities. Hence, to skip this issue, some existing approaches simply align the features of text and audio, and then feed audio as a replacement for text into text-to-image models to generate images. However, due to the lack of detailed audio-text annotations, most existing models align audio with ground truth categories, which limits more fine-grained image generation. Meanwhile, this kind of method neglects the description gap between audio and image. 

Besides, to perform end-to-end training, other researchers try to intercept images from videos to obtain image-audio pairs as training data. Nevertheless, this approach brings two problems: first, although including sufficient audio types, the quality of the obtained images cannot be ensured due to noisy source videos of traditional datasets represented by AudioSet and VGGSound. The low-quality image data will affect the performance of text-to-image models during finetuning; even the strongest open-source models will suffer severe quality degradation. Second, it is difficult to ensure that the image obtained by randomly intercepting a certain frame in the video is completely aligned with the audio, because the sounding entity in the audio does not necessarily appear in this frame at the same time. Some studies \cite{sung2023sound, zhou2025macs} are aware of this issue and choose to extract sounding objects via visual information through specific models, but such models are not general-purpose and perform poorly in various domains of wild data.

To break the current predicament in A2I, one direct way is to propose a high-quality audio-image pairing dataset with fine-grained captions. To achieve this goal, we developed a multi-stage, object-level audio-visual aligned pipeline to extract an image from a video source. However, we found that only a small number of frames can meet such high standards. Thus we further use the newly released powerful open-source model FLUX.1-Krea-dev \cite{flux1kreadev2025} to generate realistic images. Since the research benefits brought by realistic images are greater than those from data of other styles, with limited computing resources, we first consider building a set of images that are close to real photos and have a high degree of aesthetic appeal, rather than a general dataset. We name our dataset as A2I-Set, which consists of a total of 241 hierarchical classes of typical audio, pair-wise images, as well as fine-grained multimodal captions. To conclude, our main contribution can be summary as follows:
\begin{itemize}
\item We propose a large-scale dataset A2I-Set for audio-visual cross-modal research, which includes more than 200 common types of audio, including music, speech, and natural sounds. In addition, to boost A2I evaluation, we delicately label a new eval set with multiple sounding objects.
\item To obtain the dataset, we propose a complete data synthesis process, which realizes the entire workflow from audio and video of the source dataset to high-quality audio-image-text tri-paired data via SOTA models, including captioning, extraction, generation, filtering, etc.
\item To demonstrate the effectiveness of our dataset, we trained an effective audio-to-image baseline model called AudioCanvas using our A2I-Set. The experimental results show that, benefiting from the high quality of A2I-Set, AudioCanvas exhibits excellent generation performance while maintaining a high degree of cross-modal alignment.

\end{itemize}

\section{Related Work}

\subsection{Audio-Image Pairwise Dataset}
Due to the lack of corresponding audio data in common image datasets, most existing methods are constructed from existing audio datasets with video source labels such as \cite{li2022learning,owens2016visually, li2021ai, lee2022sound, gemmeke2017audio, chen2020vggsound}. As two of the most famous and large-scale audio datasets, AudioSet \cite{gemmeke2017audio} as well as VGGSound \cite{chen2020vggsound} have a huge impact on the audio generation field. However, both of them mainly focus on audio quality instead of the video that is accompanied by audio. Although VGGSound has automatic detection of video-audio correspondence in addition to audio quality, video quality is still not considered the primary criterion for screening. In recent years, some studies have identified their shortcomings in annotation quality and cross-modal pairing, and have attempted to propose improvements: VGGSounder \cite{zverev2025vggsounder} analyzes the issues of label overlap, missing, and audio-visual alignment in VGGSound, and presents a re-annotated test set of VGGSound. FusionAudio-1.2M \cite{chen2025fusionaudio} extracts contextual clues from videos to assist audio captioning, reducing the hallucination problem that occurs when single audio modal models perform annotation, and provides high-quality audio annotations with visual representations for AudioSet. To construct an Audio-Image pairwise dataset, existing methods simply perform rule-based (mid-frame, random frame, etc.) extracting frames from videos \cite{chatterjee2020sound2sight,lee2022sound,tang2023any,liu2024music,biner2024sonicdiffusion} or use a pretrained model to intercept \cite{sung2023sound, zhou2025macs}. Although the latter method has improved the former, current audio-visual video parsing models and sound source localization models only perform well on in-domain data; it is difficult to ensure good extraction capabilities for a wide range of noisy video data.

\begin{figure*}[htbp]  
    \centering  
    \includegraphics[width=1.0\textwidth]{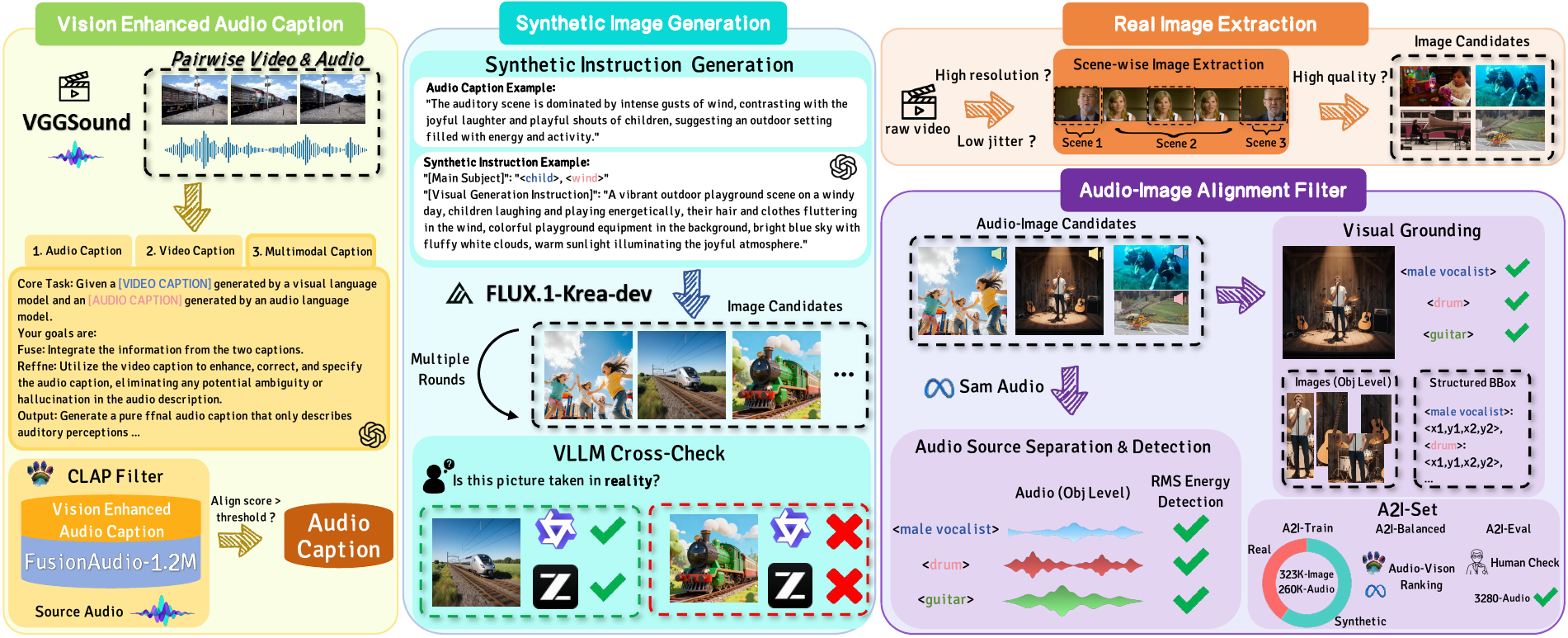}
    \caption{An overview of our proposed A2I-Set dataset synthesis pipeline, where we used multiple reliable models for captioning, filtering, generation, and evaluation. Human experts are employed to ensure data quality.}  
    \label{fig:pipeline}  
\end{figure*}

\subsection{Audio to Image Generation}
Early Audio to Image (A2I) generation mainly focused on a specific field of audio and used a small amount of data to train models from scratch \cite{chatterjee2020sound2sight}. Benefiting from the rapid development of text-to-image generation, researchers have begun to try to use audio instead of text to generate images. Among these, Generative Adversarial Networks (GANs) and Stable Diffusion models \cite{rombach2022high} are representative: Sound2Scene \cite{sung2023sound} aligns images and audio into a consistent space to achieve image-to-image generation based on ICGAN \cite{casanova2021instance}. Facing the fact that there's no available audio-image pairing dataset, AudioToken \cite{yariv2023audiotoken}, GlueGen \cite{qin2023gluegen}, and SonicDiffusion \cite{biner2024sonicdiffusion} choose to directly train a model to convert audio into text tokens and use them as conditional prompts for diffusion models to drive image generation. More recently, SoundAdapter \cite{wang2025draw} further proposes a multi-granularity method to achieve better alignment between audio and clip embedding. MACS \cite{zhou2025macs} proposes a two-stage solution which separates audio first, then performs IP-Adapter \cite{ye2023ip} like end-to-end training based on a diffusion model. However, this approach relies heavily on a well-developed audio-image dataset.

\subsection{Synthetic Dataset}
Benefiting from the rapid development of generative models, many works have begun to synthesize data as training data in recent years \cite{ye2025echo,chen2025blip3,gu2025realsyn}. Synthetic data can play as a supplement to rare scenarios in real-world datasets, providing clear and controllable supervision\cite{ye2025echo}. BLIP3-o \cite{chen2025blip3} leverages approximately 10k image-text pairs generated by GPT-4o during fine-tuning, which yields significant improvements in instruction-following performance. RealSyn \cite{gu2025realsyn} designs a synthetic text generation module to construct datasets and promote vision-language representation learning. In speech generation \cite{9003990,liu25e_interspeech}, using synthetic data for training is a common and natural choice. In our task, we find that after filtering nearly one million candidate frames from the video source, we only obtained approximately 100,000 valid images. This is because many audio elements themselves are difficult to appear in the same frame simultaneously. To better achieve universal audio-to-image generation, we used FLUX.1-Krea-dev \cite{flux1kreadev2025} to simulate real images.

\section{Dataset}
In this section, to ensure data quality, we construct a multi-stage pipeline which includes captioning, extraction, generation and filtering stages, building upon two typical audio datasets. We build a large-scale audio-image dataset through multiple strong models, namely A2I-Set. The constructed dataset is characterized by its large size, high quality, and high audio-visual matching. It encompasses a wide range of audio categories, including music, natural sounds, and speech. This comprehensive and diverse dataset is designed to promote cross-modal research between audio and images.

\begin{figure*}[htbp]
    \centering
    \begin{minipage}[t]{0.3\linewidth}
        \centering
        \includegraphics[width=\linewidth]{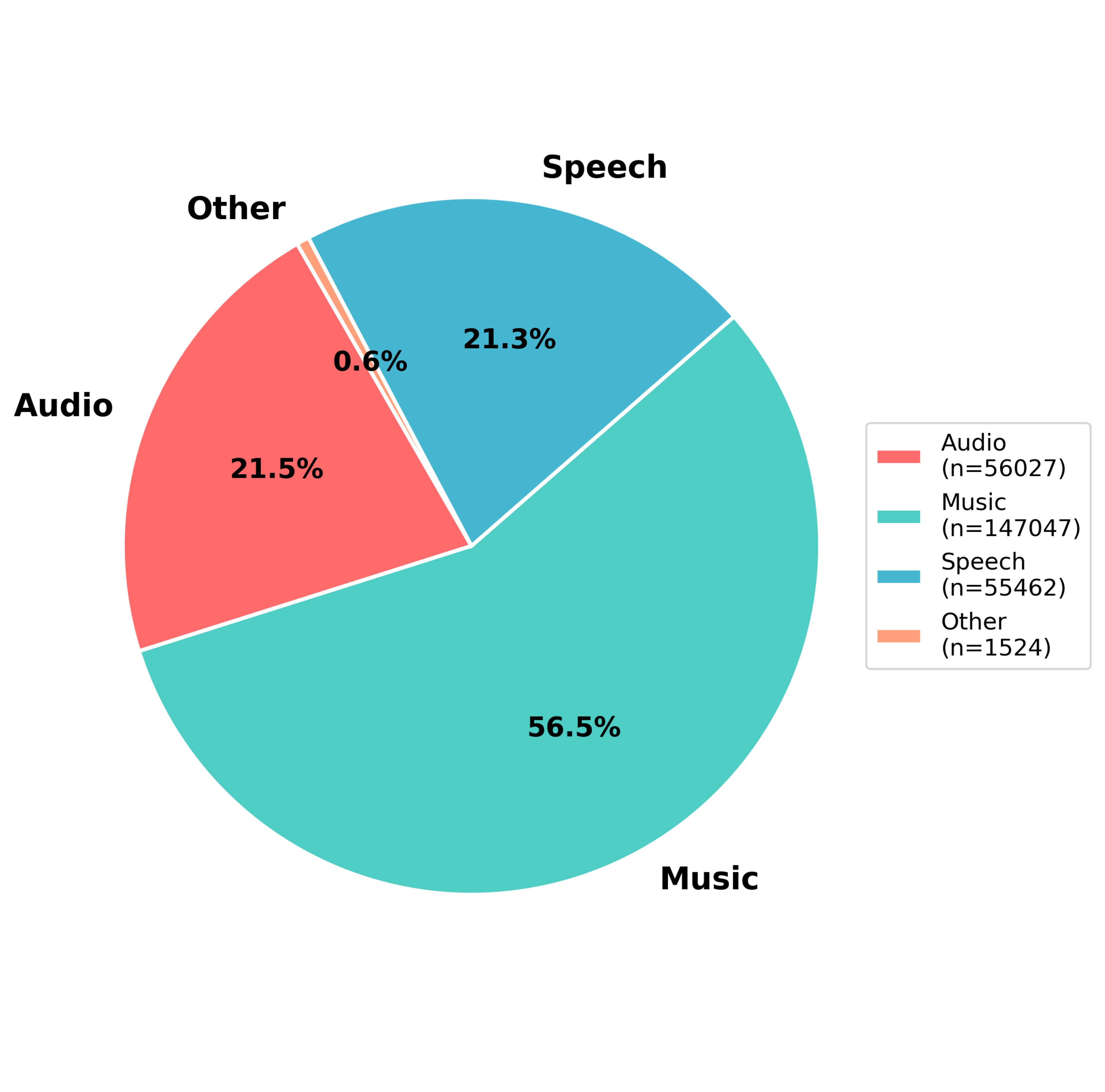}  
        \caption{The proportion of different categories of audio in the training set of A2I-Set.}
        \label{fig:classes}
    \end{minipage}
    \hfill  
    \begin{minipage}[t]{0.3\linewidth}
        \centering
        \includegraphics[width=\linewidth]{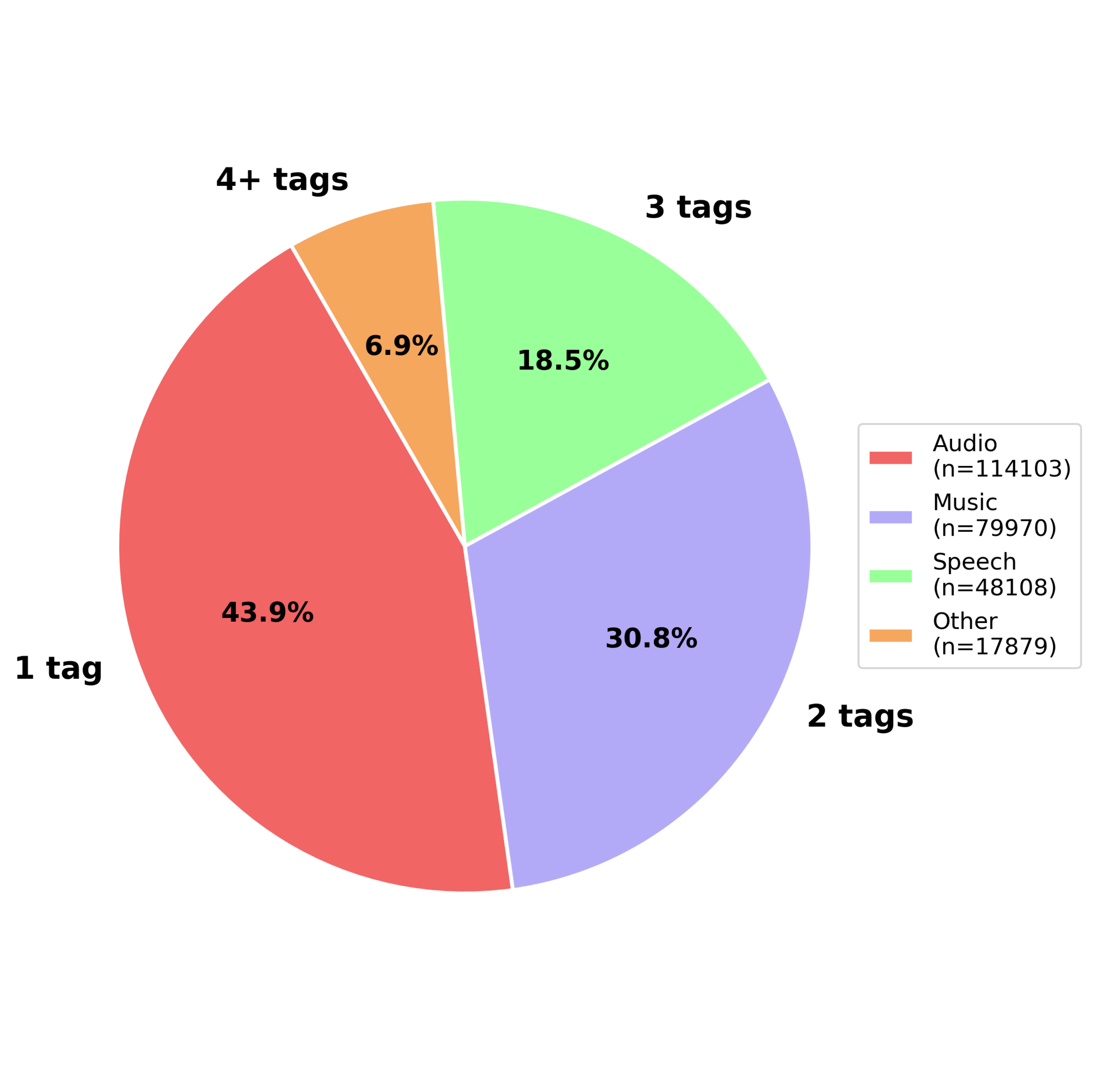}
        \caption{Proportion of the number of sound sources in the audio of A2I-Set. A2I-Set has a uniform proportion of sound sources.}
        \label{fig:tags}
    \end{minipage}
    \hfill  
    \begin{minipage}[t]{0.3\linewidth}  
        \centering  
        \includegraphics[width=\linewidth]{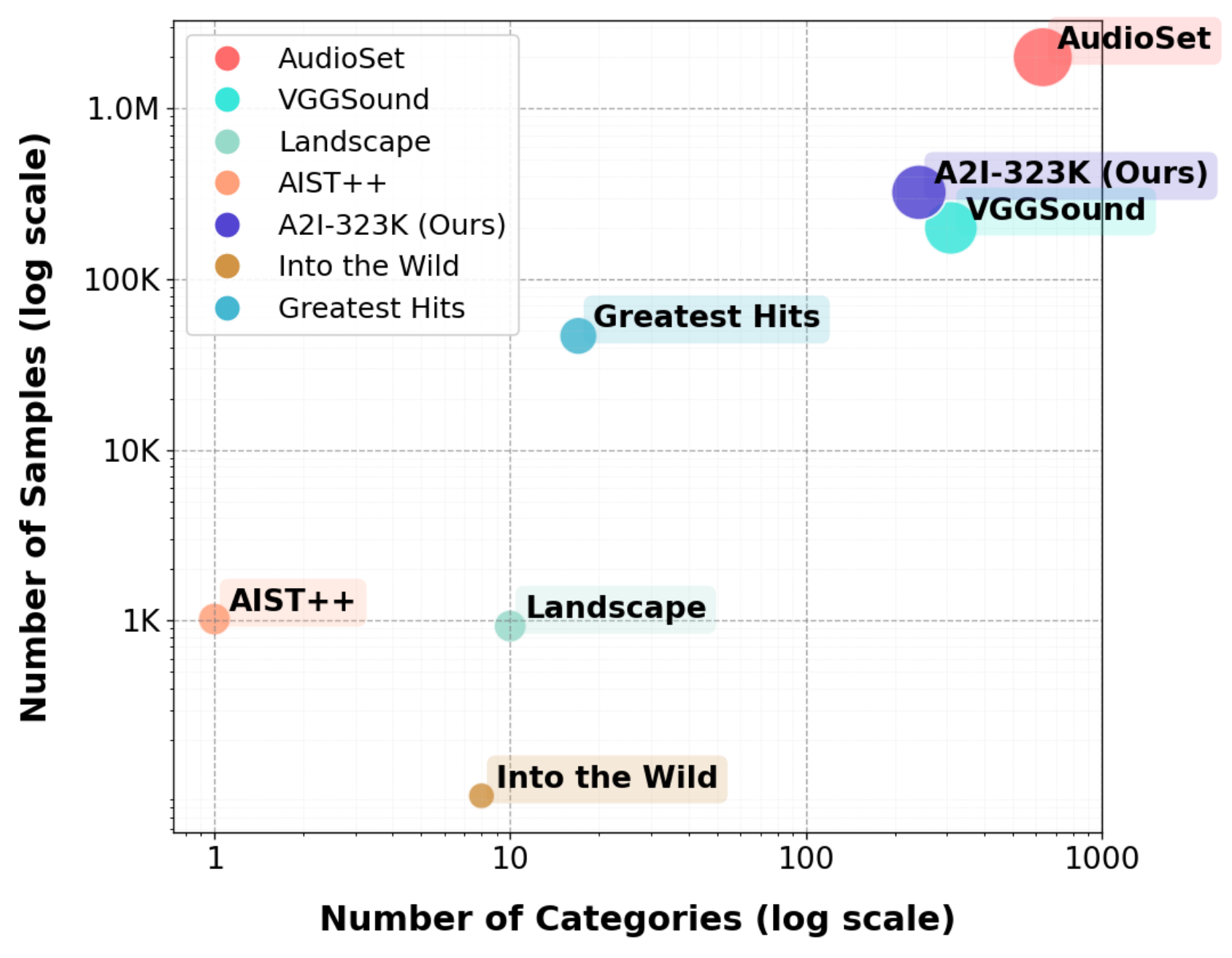}
        \caption{Comparison of A2I-Set with several other audio-visual related datasets.}  
        \label{fig:dataset}  
    \end{minipage}
\end{figure*}
\subsection{Vision Enhanced Audio Caption}
In order to quickly obtain audio and video with a high degree of matching, we directly utilize two existing datasets: AudioSet consists of 632 hierarchically inclusive audio event classes and a collection of around 2M human-labelled 10-second sound clips drawn from YouTube videos. VGGSound contains approximately 200k video clips, covering 309 audio categories, with a total duration of about 550 hours. Each clip is fixed at 10 seconds, while the matching between audio and source video has been simply checked. Thanks to FusionAudio-1.2M \cite{chen2025fusionaudio}, which has already incorporated video information to caption the audio from AudioSet, we directly incorporate this part of the captions into our collection. Meanwhile, we processed the VGGSound dataset in a similar manner. Due to the long time since AudioSet and VGGSound datasets were released, some audio or videos are no longer available. Eventually, we obtained a total of around 1.1M of valid audio entries from these sources. 

Like FusionAudio-1.2M \cite{chen2025fusionaudio} does to AudioSet, we use the video caption model PLLaVA \cite{xu2024pllava} and the audio caption model Qwen-audio-2 \cite{chu2024qwen2} to respectively obtain the information from pairwise source videos and audio data in VGGSound and integrate the two through the GPT-4o-mini model to yield a more reliable audio caption with potential vision clues. This step is because a single audio caption model always suffers from severe hallucination problems, which will affect the quality of the data. The introduction of visual information has greatly alleviated this phenomenon. At the same time, doing so allows our data to be aligned with FusionAudio-1.2M, enabling unified subsequent processing. Since the correspondence between audio and video in the source data cannot be perfectly guaranteed, to obtain reliable audio captions, we further filter all generated audio captions through CLAP \cite{laionclap2023}, a large-scale audio-text cross-modal contrastive learning pre-trained model. By calculating the similarity scores of the two modalities, we filtered out nearly 40\% of the total data.

\subsection{Real Image Extraction}

To obtain high-quality images, we first filtered high-resolution videos. Specifically, videos need to be no less than 540$\times$360. Additionally, we use an optical flow estimation method \cite{farneback2003two} to perform jitter filtering on the videos to obtain stable videos for extracting images. Next, we utilize TransNet V2 \cite{soucek2020transnetv2} to perform scene segmentation on videos and extract the middle frame for each scene. To ensure image quality, we further used the image quality assessment model ARNIQA \cite{agnolucci2024arniqa} for evaluation and filtered out images with a score lower than 0.4. All qualified images are taken as image candidates and will be evaluated for audio-image alignment in subsequent steps.

\subsection{Synthetic Image Generation}

In addition to extracting real image alternatives, we also used FLUX.1-Krea-dev to generate images. This is because, due to the temporal nature and mixed characteristics of audio, it is difficult for a single frame image in a video to be completely aligned with the source audio. After obtaining reliable audio captions in the first stage, we find that simply inputting descriptions from the audio modality into the image generation model yielded poor performance. Obviously, this is because the audio description is not completely aligned with the input of the text-to-image model. Therefore, we used GPT-4o-mini to adjust the audio descriptions again. Unlike previous instructions, at this stage, we allowed the model to appropriately expand the audio captions by incorporating potential visual descriptions within them, and to construct reliable drawing instructions from a professional perspective, including elements such as subject identification, visual descriptions, atmosphere recognition, colors, and composition. Detailed and professional image generation instructions will be uniformly input together with clear instructions for generating realistic style images into FLUX.1-Krea-dev \cite{flux1kreadev2025}, a newly launched powerful image generation model characterized by realism. However, the quality of image generation remains uncontrollable, and the model still often generates unrealistic-style images. At the same time, phenomena such as blurriness and distortion can occasionally be observed in the generated images. We used two of the most powerful open-source visual large models (VLMS): Qwen2.5-VL-72B-Instruct \cite{qwen2.5-VL} and GLM-4.5V \cite{hong2025glm} to visually screen the generated images. We have made various attempts with visual prompts. Finally, we find that directly having the model judge whether the images generated by FLUX.1-Krea-dev are real photos achieved the best results in judging the model's style and visual details. For detailed instructions on VLMs, please refer to our supplementary material.

Only synthetic images that are judged as qualified by both visual models can pass the screening. The overall pass rate is only 10\% to 20\%. To generate a sufficient number of images, we conducted multiple rounds of generation. In this stage, we finally obtain approximately 240K qualified with a high resolution of 1024$\times$1024 images, which will be filtered together with real image candidates in the next stage.

\subsection{Audio-Image Alignment Filter}

As mentioned earlier, the frames randomly captured from the video may not fully correspond to the elements in the audio. At the same time, large models may also encounter hallucinations when judging audio elements. Therefore, to ensure the rigour of the dataset, we used Sam Audio \cite{shi2025sam} to perform object-level separation of audio through prompts. By performing RMS energy detection on the separation results, we ensure that the object elements are present in the audio. Next, to judge the existence of object elements, we performed visual grounding separately for each object element using Qwen3-VL-Flash \cite{bai2025qwen3} and Ground Dino \cite{liu2023grounding}. In this way, we obtained highly aligned audio-image pairs, and a detailed analysis will be provided in the next section.

\subsection{Data Analysis}
A2I-Set contains 323,446 images and 260,060 audios with 241 hierarchical labels, which means an audio may be paired with multiple valid images. Specifically, A2I-Set includes 193,653 synthetic images and 129,773 real images. In particular, to encourage audio-visual research in multi-source scenes, we also select 3,280 evaluation image-audio pairing data as evaluation dataset A2I-eval from the same source data obtained from our data construction pipeline, where the audio categories are the same as the training set. The final evaluation set is manually screened by field experts to ensure audio object correction for element-wise evaluation. In addition, we notice that the long-tail problem exists in A2I-Set, so we further propose an A2I-balanced set. A2I-balanced is a 10K sample selected from the synthetic images in A2I-Set by ranking both ImageBind scores and aesthetic scores, and also has undergone expert evaluation to ensure visual quality. We provide some downstream applications based on A2I-balanced in the appendix to further highlight the high quality of the dataset.

\cref{fig:classes} shows the distribution of audio types in A2I-Set, highlighting its universal diversity, which contrasts with small specialized datasets such as the natural sound dataset Landscape \cite{lee2022sound}, Into the Wild \cite{li2022learning}, Greatest Hits\cite{owens2016visually}, and the dance music dataset AIST++ \cite{li2021ai}. Such diversity secures our dataset with an abundant variety and general knowledge, which can be applied in model training. In addition, we have also counted the distribution of the number of sound sources in the audio. As can be seen from \cref{fig:tags}, our dataset is balanced across various numbers of sound sources, indicating that our dataset provides both simple (single sound source) and complex (multiple sound sources) training data. Meanwhile, our dataset is consistent with the view of MACS \cite{zhou2025macs}, real-world audio is mainly mixed audio rather than single-source audio. This further underscores our purpose, which is to construct a dataset that can better mimic the real world. As shown in \cref{fig:dataset}, we compare it with several related multimodal datasets, demonstrating A2I-Set's advantage in data scale and the number of audio classes.

\section{Method}

\subsection{Problem Formulation}
Let an audio signal $a \in  \mathcal{A}$ correspond to a visual scene $x \in  \mathcal{X}$. Our goal is to learn a generative model $G$ capable of synthesizing an image $\hat{x}$ conditioned solely on the audio input:
\begin{equation}
  \hat{x} = G(a).
  \label{eq:1}
\end{equation}

To train such a neural network $G$, a dataset containing audio-visual data pairs $\mathcal{D} = \{X_i, A_i\}_{i=1}^N$ is generally required. However, directly training such a general modal alignment model requires a fairly large number $N$ of training data, which is usually difficult to achieve. Therefore, we utilize the image generation capabilities of an existing pretrained generation model, specifically text-to-image (t2i) diffusion-based generator $\hat{G}$ conditioned on text embeddings $e_t$. To obtain such text embedding, a text encoder $E_t \in \mathbb{R}^{d_t}$ is typically needed to encode the input text $y$ into latent space. Therefore, our problem is transformed into how to embed audio embeddings $e_a$ into the pre-trained t2i model. Most previous methods sidestep this challenge by directly mapping audio features to text features. Several years ago, image-to-image (I2I) generation also faced similar challenges. IP-Adapter \cite{ye2023ip} proposed a decoupled cross-attention module, which was originally designed as an adapter to enable a pretrained text-to-image diffusion model to generate images with an image prompt. Hence, most works in A2I, such as MACS \cite{zhou2025macs}, use it for the injection of audio embeddings. Notably, we propose a new injection method named FiLM-weighted Audio-Text Fusion, which will be introduced later.

\begin{figure*}[!t]  
    \centering  
    \includegraphics[width=1.0\textwidth]{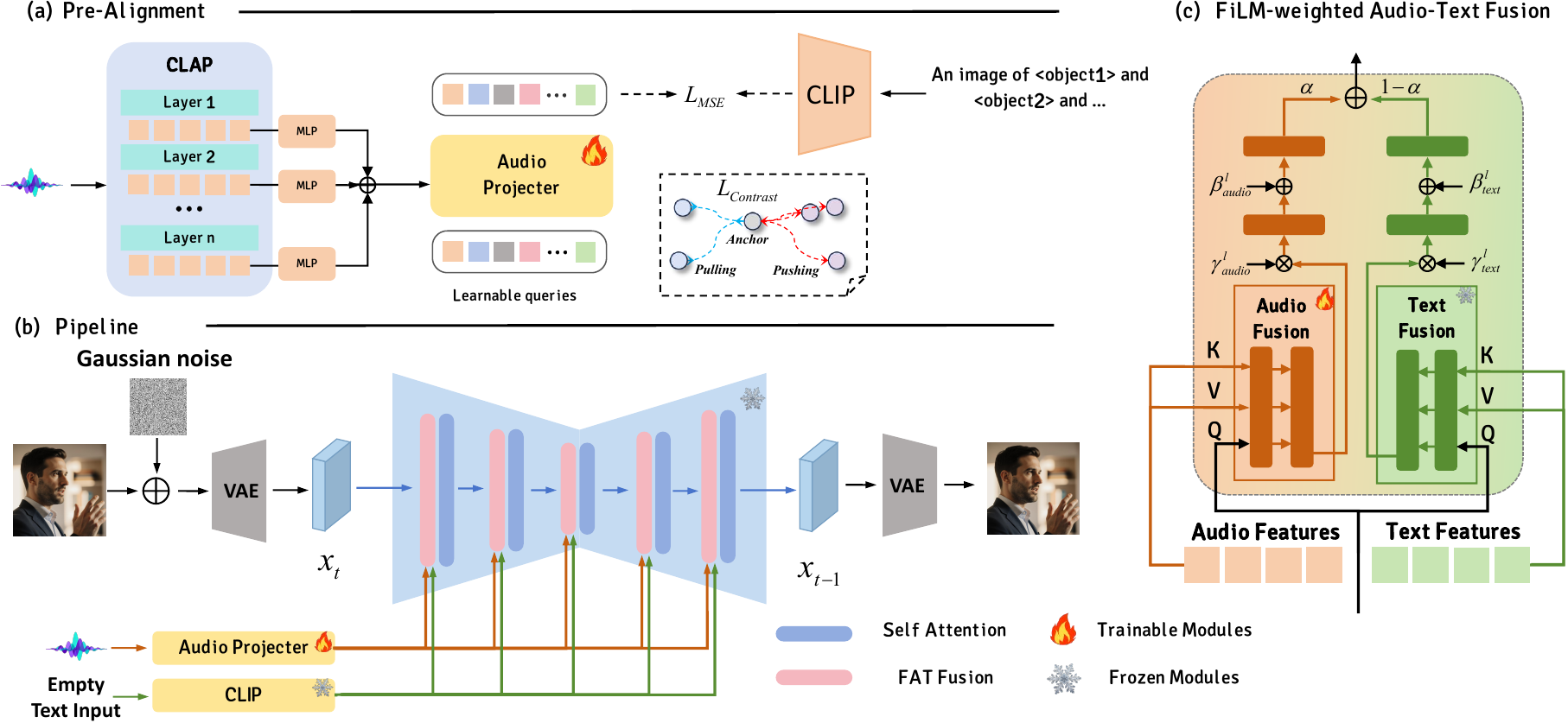}
    \caption{The overview of audio-to-image generation architecture AudioCanvas, audio features are injected into the diffusion model through FAT-Fusion. We keep the text input empty for both training and inference stages. Only few modules need training while the diffusion model backbone is frozen.}  
    \label{fig:ipadapter}  
\end{figure*}

\subsection{Pre-Aligned Audio Projector}
 When training a stable diffusion model (here we utilize SD 1.4), a large dataset of text-image pairs is typically used to learn a good text-image space correspondence. To preserve this good alignment as much as possible and achieve better performance, we first introduce an Audio Projector module with Pre-Alignment before fine-tuning the Stable Diffusion model. As \cref{fig:ipadapter} (a) shows, we extract multi-level features of the CLAP model, adjust the dimensions using MLP, then concatenate and input them together into the Audio Projector. The Audio Projector extracts useful information from the CLAP features through learnable queries and attempts to align with the CLIP model \cite{radford2021learning} output. Thanks to the object elements corresponding to the audio in our annotations, we directly use "An image of <object1> and <object2> ..." as the input to CLIP.

To enable audio conditioning, we first adopt the pretrained audio encoder $E_a (\cdot)$ from CLAP \cite{laionclap2023} that extracts multi-layer audio embeddings $e_{a1}, e_{a2}, e_{a3}, e_{a4} = E_a(a)$ from audio signals. Then each audio embedding is projected by an MLP separately to obtain a uniform hidden dimension $\mathbb{R}^{d_a}$. $e_{a1}$, $e_{a2}$, $e_{a3}$, and $e_{a4}$ are then concatenated and input into the Audio Projector, finally yielding the aggregated audio representation $e$. To better align the output of the Audio Projector with the CLIP space, we use a joint-training strategy that combines Mean Squared Error (MSE) loss and contrastive loss functions. Specifically, we used the InfoNCE \cite{alayrac2020self} loss, treating audio clips from the same class as positive pairs and audio clips from different classes as negative pairs. Given a batch of audio features input $A = \{e_1,\ldots,e_N\}$, the InfoNCE loss can be denoted as:

\begin{equation}
   \mathcal{L}_{\text{InfoNCE}} = -\log \frac{ \exp\left(\langle \mathbf{e}_i, \mathbf{e}_{pos} \rangle\right)}{\sum_{neg \in Neg} \exp\left(\langle \mathbf{e}_i, \mathbf{e}_{neg} \rangle\right)}.
  \label{eq:2}
\end{equation}

where ${A}_{pos}$ shares same object between $e_i$, while ${e}_{neg}$ in negative set $Neg$ don't. $\langle \cdot \rangle$ denotes inner product between elements. Besides, the MSE loss is defined as the squared norm difference between the audio feature $a_i$ and the text feature $t_i$:

\begin{equation}
\mathcal{L}_{\text{MSE}} = \left\| a_i - t_i \right\|_2^2.
  \label{eq:3}
\end{equation}

The total training object $\mathcal{L}_{\text{Total}}$ of audio projector combines these two losses:

\begin{equation}
\mathcal{L}_{\text{Total}} = \mathcal{L}_{\text{MSE}} + 0.2*\mathcal{L}_{\text{InfoNCE}}.
  \label{eq:4}
\end{equation}

\subsection{FiLM-weighted Audio-Text Fusion}

The projected audio feature $e$ obtained from Audio Projector is then integrated into the pre-trained UNet model by the \textbf{F}iLM-weighted \textbf{A}udio-\textbf{T}ext Fusion (FAT Fusion) module. Among these, the cross-attention layers used for text features and image features are separated. Specifically, we add a new cross-attention layer to each cross-attention layer of the original UNet model to insert audio features, which can be denoted as: 

\begin{equation}
    \mathbf{Z}_a = \text{Attention}(\mathbf{Q}, \mathbf{K_a}, \mathbf{V_a}) = \text{Softmax}\left( \frac{\mathbf{Q}\mathbf{K_a}^\top}{\sqrt{d_a}} \right) \mathbf{V_a},
  \label{eq:5}
\end{equation}

where $\mathbf{Q}=\mathbf{Z}\mathbf{W_q}$ is the query matrices from UNet latent, $\mathbf{K_a}=\mathbf{e'}\mathbf{W_k}$, $\mathbf{V_a}=\mathbf{e'}\mathbf{W_v}$ are key and values matrices computed from the audio features through corresponding weight matrices $\mathbf{W_ka}$ and $\mathbf{W_va}$. $d_a$ denotes the dimension of learned features $e'$.

In addition to audio conditions, the model also accepts text condition injection through cross-attention. Therefore, for each layer of UNet, the latent features undergo two updates. Specifically, the features integrated with text information and the features integrated with audio information are summed to serve as the output latent. Thus, the final formulation of the decoupled cross-attention is defined as follows:

\begin{equation}
    \mathbf{Z}_t = \text{Attention}(\mathbf{Q}, \mathbf{K_t}, \mathbf{V_t}) = \text{Softmax}\left( \frac{\mathbf{Q}\mathbf{K_t}^\top}{\sqrt{d_t}} \right) \mathbf{V_t},
  \label{eq:6}
\end{equation}

where $\mathbf{K_a}=\mathbf{e_t}\mathbf{W_k}$, $\mathbf{V_a}=\mathbf{e_t}\mathbf{W_v}$. Notably, in both cross-attention, the query $\mathbf{Q}$ remains the same. After obtaining the text representation $\mathbf{Z}_t$ and the audio representation $\mathbf{Z}_a$, instead of directly summing them, we propose a flexible feature fusion technique similar to Feature-wise Linear Modulation (FiLM)\cite{perez2018film}. FiLM is a feature-level linear modulation technology that performs Affine Transformation on the intermediate features of the neural network, enabling the model to dynamically adjust feature expressions according to inputs. 

Specifically, as \cref{fig:ipadapter} shows, given representations $\mathbf{Z}_a$ and $\mathbf{Z}_t$ in the $lth$ layer of unet, each representation is modulated by two channel-wise feature-aware modulation parameters $\gamma^l, \beta^l \in \mathbb{R}^{B \times 1 \times D}$. In practice, we adopt $D=d_a=d_t$. This process can be denoted as:

\begin{align}
\gamma_{\text{audio}}^l = f_{audio}(\mathbf{Z}_a^l), \gamma_{\text{text}}^l = f_{text}(\mathbf{Z}_t^l), 
\\
\beta_{\text{audio}}^l = h_{audio}(\mathbf{Z}_a^l), \beta_{\text{text}}^l = h_{text}(\mathbf{Z}_t^l),
\end{align}

\begin{align}
\mathbf{Z}_{a'}^l = \text{FAT-Fusion}_{\text{audio}}(\mathbf{Z}_a^l) &= (1+\gamma_{\text{audio}}^l) \cdot \mathbf{Z}_{a}^l + \beta_{\text{audio}}^l, \\
\mathbf{Z}_{t'}^l = \text{FAT-Fusion}_{\text{text}}(\mathbf{Z}_t^l) &= (1+\gamma_{\text{text}}^l) \cdot \mathbf{Z}_{t}^l + \beta_{\text{text}}^l,
\end{align}

\begin{equation}
\mathbf{Z}_{new}^l = (1-\alpha)\mathbf{Z}^l_{t'} + \alpha\mathbf{Z}^l_{a'}.
  \label{eq:7}
\end{equation}

In practice, we keep $\alpha=0.5$ to ensure modal balance. Since the text information corresponding to the audio is generally unknown during inference, we keep the input text empty for extraction during both training and inference.

\subsection{Training and Inference}

During the training process, we finetune AudioCanvas while keeping the parameters of the pre-trained diffusion model unchanged. AudioCanvas is trained on our proposed paired audio-image dataset A2I-Set, using the same training objective as the original stable diffusion:

\begin{equation}
L_{SD} = \mathbb{E}_{{x}_0, {\epsilon}, y, a, t} \|{\epsilon} - {\epsilon}_\theta({x}_t, y, a, t)\|^2.
  \label{eq:8}
\end{equation}

In addition, to ensure style consistency, we also set a probability of $p=0.1$ to randomly drop the audio conditions. Since the text condition has always been set to empty, the model performs unconditional generation at this point, which also enables classifier-free guidance in the inference stage.

\begin{table*}[htbp]
  \centering
  \caption{Performance comparison on A2I-eval (mixed-source). The best results are \textbf{bold}, and the second-best results are \underline{underlined}. Training data refers to the estimated additional training data related to visual modality and audio modality introduced during the model training process. $^\dagger$ denotes the model designed for sound generation (instead of general audio).}
  \label{tab:performance1}
  \setlength{\tabcolsep}{6pt}
  \begin{tabular}{ccccccccc}
    \toprule
    Method & $\text{FID}\downarrow$ & $\text{IS}\uparrow$ & $\text{Aes}\uparrow$ & $\text{HPSv3}\uparrow$ &  $\text{AIS}\uparrow$ & $\text{TIS}\uparrow$ & Backbone & training data \\
    \midrule
    AudioToken \cite{yariv2023audiotoken} & \textbf{69.79} & 24.52 & \underline{5.42} & 0.2168 & 0.0721 & 0.2395 & SD1.4 & $\approx 200$k\\  
    Sound2Scene $^\dagger$ \cite{sung2023sound} & 79.95 & 16.54 & 3.85 & -5.8154 & 0.0712 & 0.2380 & BigGAN & $< 100$k\\  
    SonicDiffusion $^\dagger$ \cite{biner2024sonicdiffusion} & 91.57 & 12.55 & 4.94 & -2.2509 & 0.0468 & 0.2043 & SD1.4 & $\approx 22$k\\
    SoundAdapter \cite{wang2025draw} & 74.55 & 21.51 & 5.04 & -1.3218 & 0.0613 & 0.2329 & SD1.5 & $\approx 200$k\\
    CoDi \cite{tang2023any}  & 75.89 & \underline{24.75} & 5.41 & \underline{2.9352} & \underline{0.0837} & \underline{0.2489} & SD1.5 & $\approx 100$M \\  
    MACS \cite{zhou2025macs} & \underline{72.27} & 22.49 & 4.67 & 0.4623 & 0.0792  & 0.2189 & SD1.4 & $\approx 100$k\\  
    \midrule
    \textbf{AudioCanvas} (ours) & 73.93 & \textbf{26.07} & \textbf{5.47} & \textbf{4.8115} & \textbf{0.0841} & \textbf{0.2547} & SD1.4 & $\approx 323$k \\  
    \bottomrule
  \end{tabular}
\end{table*}

\begin{table*}[htbp]
  \centering
  \caption{Performance comparison on Landscape (single-source). The best results are \textbf{bold}, and the second-best results are \underline{underlined}. Training data refers to the estimated additional training data related to visual modality and audio modality introduced during the model training process. $^\dagger$ denotes the model designed for sound generation (instead of general audio).}
  \label{tab:performance2}
  \setlength{\tabcolsep}{6pt}
  \begin{tabular}{ccccccccc}
    \toprule
    Method & $\text{FID}\downarrow$ & $\text{IS}\uparrow$ & $\text{Aes}\uparrow$ & $\text{HPSv3}\uparrow$ &  $\text{AIS}\uparrow$ & $\text{TIS}\uparrow$ & Backbone & training data \\
    \midrule
    AudioToken \cite{yariv2023audiotoken} & \textbf{100.42} & \textbf{15.66} & 5.50 & 1.9184 & 0.0718 & 0.2339 & SD1.4 & $\approx 200$k\\  
    Sound2Scene $^\dagger$ \cite{sung2023sound} & 120.86 & 7.98 & 4.18 & -1.7713 & \textbf{0.1039} & 0.2485 & BigGAN & $< 100$k\\  
    SonicDiffusion $^\dagger$ \cite{biner2024sonicdiffusion} & \underline{110.05} & 10.26 & 5.14 & 3.3671 & 0.0844 & 0.2370 & SD1.4 & $\approx 22$k\\
    SoundAdapter \cite{wang2025draw} & 127.54 & \underline{11.83} & 5.11 & 2.3801 & 0.0866 & \textbf{0.2628} & SD1.5 & $\approx 200$k\\
    CoDi \cite{tang2023any}  & 154.56 & 6.93 & \underline{5.59} & \underline{5.6715} & \underline{0.1031} & 0.2585 & SD1.5 & $\approx 100$M \\  
    MACS \cite{zhou2025macs} & 124.62 & 9.05 & 4.59 & 0.6207 & 0.0786 & 0.2364 & SD1.4 & $\approx 100$k\\  
    \midrule
    \textbf{AudioCanvas} (ours) & 137.03 & 7.31 & \textbf{5.75} & \textbf{5.8728} & 0.0952 & \underline{0.2592} & SD1.4 & $\approx 323$k \\  
    \bottomrule
  \end{tabular}
\end{table*}

\section{Experiments}

\subsection{Implementation details}
We train AudioCanvas on our proposed A2I-Set. During training, we keep the text input empty to prevent any additional text from affecting the model. To ensure the model to learn strong high-level aesthetic priors, we first train it for 25 epochs on the synthetic part of the A2I-Set, followed by 45 additional epochs on the entire dataset. All training was conducted on a single H100 machine with 8 cards. During training, a probability of $p=0.1$ is used to randomly drop the audio conditions to ensure the style transfer of the model, which enables our model to generate stably facing missing or highly noisy inputs. The training stage utilizes the AdamW optimizer with a batch size of 8 samples per GPU. We also use gradient accumulation with a step of 8. The hyperparameters of optimizer are set to $\beta_1 = 0.9$, $\beta_2 = 0.99$, with a weight decay of 0.01. In terms of the selection of generative models, we have observed that most audio-to-image models adopt early versions of SD. To ensure a fair comparison, we have also followed this selection and used SD v1.4 as our image generation backbone. We further provide more experiments and qualitative analysis using stronger backbones (SD2 and SD-XL) in the Appendix.

During the inference phase, to ensure the comprehensiveness of the results, we used two datasets to test our model. A2I-eval contains 3,280 manually-label audio-image data pairs with all 241 classes in A2I-Set. In addition, we also use the single-source dataset Landscape \cite{lee2022sound}, which contains 9 distinct labels for 1,000 video clips, as an exogenous single-source evaluation dataset. Due to the small size of the Landscape dataset, to ensure the stability of the test, we generate multiple images for each audio during inference. For AudioCanvas, we choose cfg scale of 7.5 for A2I-eval and 8.5 for Landscape.

\subsection{Evaluation Metrics}
We use multiple metrics to evaluate our model, which can be divided into three parts: 1) Visual quality of the generated images, which includes Fréchet Inception Distance (FID) \cite{heusel2017gans}, and Inception Score (IS) \cite{salimans2016improved}. 2) Human preference of the generated images, represented by Aesthetic evaluation score (Aes) and HPSv3 score \cite{ma2025hpsv3}. Aes is directly extracted by the LAION-Aesthetics Predictor V2. Notably, HPSv3 requires a text prompt during inference; we directly input a template sentence filled with ground truth labels. 3) Modality alignment, following \cite{yariv2023audiotoken}, we introduce Audio-Image Similarity (AIS), which assesses the similarity between the source audio and the generated images using the Wav2CLIP model \cite{wu2022wav2clip}. In addition, we do not adopt the Image-Image Similarity (IIS) proposed in \cite{yariv2023audiotoken}, as the ground truth image of audio is a vague concept. Instead, we directly measure the similarity between the generated image and the ground truth labels with CLIP \cite{radford2021learning}, namely Text-Image Similarity (TIS). In addition, we also indicate the backbones and the amount of training data used by different methods in \cref{tab:performance1} and \cref{tab:performance2}, because the generative ability of the model is closely related to them.
\subsection{Main Results}
In \cref{tab:performance1} and \cref{tab:performance2}, we compare several representative models \cite{yariv2023audiotoken,sung2023sound,tang2023any,zhou2025macs,wang2025draw,biner2024sonicdiffusion}. Noted that we do not select recent integrated foundation models for comparison, out of concern that the disparities in model scale and volume of training data would lead to unfair comparisons. As seen from the tables, AudioCanvas achieves the most balanced performance between visual quality and modal alignment. AudioCanvas performs excellently in all human preference scores on both test datasets, and IS on A2I-eval. Notably, our model achieves better performance than CoDi, which has approximately 100M visual training data. This can be attributed to the powerful visual quality of A2I-Set; Similar findings were also obtained in \cite{dai2023emu}, where significantly improving the quality of the denoising model can be achieved by performing quality-tuning on it with a small amount of high-quality data. More details about model generation performance and comparisons with baseline models can be find in the Appendix.

At the same time, AudioCanvas performs excellently in terms of audio-visual relevance. We attribute this to two points: 1. A2I-Set contains a large amount of multi-source data, which has great learning value for the model. Although the AudioCanvas does not specially process the multi-source data, it can also master the corresponding relationships through extensive learning. 2. Thanks to the delicate pipeline, the high-quality of A2I-Set cross-modal pairings can bring more efficient training, even surpassing 100M of rough matching pairs in CoDi. Besides, AudioToken, which achieves excellent visual-generation performance by keeping a completely frozen backbone, performs poorly in terms of audio-visual relevance. This may be because AudioToken uses indirect text tokens rather than audio as input to the diffusion model, which reveals the importance of end-to-end training.

\begin{table}[htbp]
  \centering
  \small
  \caption{Comparison of AudioCanvas on different experiment settings or model components in A2I-eval. The best results are \textbf{bold} and the second-best results are \underline{underlined}.}
  \setlength{\tabcolsep}{6pt}
  \begin{tabular}{ccccccc}
    \toprule
    \multicolumn{1}{c}{Method} &  \multicolumn{1}{c}{FID$\downarrow$} &  \multicolumn{1}{c}{IS$\uparrow$} &  \multicolumn{1}{c}{Aes$\uparrow$} & \multicolumn{1}{c}{HPSv3$\uparrow$} &  \multicolumn{1}{c}{AIS$\uparrow$} &  \multicolumn{1}{c}{TIS$\uparrow$} \\
    \midrule
    \multicolumn{1}{c}{Real-Only} & \textbf{69.86} & \underline{25.47} & 4.63 & 2.0524 & \textbf{0.0948} & \underline{0.2532} \\
    \multicolumn{1}{c}{Synth-Only} & 91.65 & 21.96 & \textbf{5.65} & \textbf{5.0926} & 0.0771 & 0.2493  \\
    \multicolumn{1}{c}{Mixture} & \underline{73.93} & \textbf{26.07} & \underline{5.47} & \underline{4.8115} & \underline{0.0841} & \textbf{0.2547} \\
    \midrule
    \multicolumn{1}{c}{w FiLM-Large} & 76.29 & 24.27 & 5.39 & 4.4098 & 0.0798 & 0.2487 \\
    \multicolumn{1}{c}{w/o FiLM} & \underline{74.87} & \textbf{27.49} & \underline{5.40} & \underline{4.4528} & \underline{0.0832} & \underline{0.2532}  \\
    \multicolumn{1}{c}{AudioCanvas} & \textbf{73.93} & \underline{26.07} & \textbf{5.47} & \textbf{4.8115} & \textbf{0.0841} & \textbf{0.2547} \\
    \bottomrule
  \end{tabular}
  \label{tab:ablation_audio_adapter}
\end{table}

\subsection{Ablation Studies}
In \cref{tab:ablation_audio_adapter}, we first report the performance of AudioCanvas on different data splits of A2I-Set. All experiments are conducted for the same training settings. "Real-Only" and "Synth-Only" denote AudioCanvas models that are trained on real and synthetic images of A2I-Set, respectively. "Mixture" follows the original training recipe of AudioCanvas. We can see from the table that synthetic data serves as the core source for the model to learn high aesthetic features, while purely real data can only enhance the realism of generated images (higher FID). Our training strategy that mixes real and synthetic data plays a key role for AudioCanvas to achieve optimal overall performance. It greatly improves the aesthetic score of images while preserving authenticity, and also delivers outstanding performance in audio-image alignment. Besides, we further discover the effect of FiLM in FAT-Fusion. We can see from the table that the FiLM module provides significant improvements to the model's audio-image alignment and aesthetic performance. However, when we further increase the parameters with a 3-layer MLP (denoted as FiLM-Large), it leads to performance degradation. This may be because an overly deep network disrupts the favourable feature distribution. To sum up, AudioCanvas has comprehensively achieved leading overall performance on the A2I-eval dataset, verifying the rationality of its data strategy and model component design.

\section{Limitations}

We mainly evaluate our model on A2I-eval, which shares the same source data with our training data. This may bring potential impact on evaluation fairness. To amend this issue, we bring more test on Landscape. Besides, to inherit good aesthetic styles, We train our model on Flux‑Style images, which may bring possible style overfitting issue. Meanwhile, we introduce real images as training data for robust real-world generalization, yet relatively low aesthetic images reduce the artistic performance of our model. We have not fully explored the trade-off of real and synthetic data.

\section{Conclusion}
To address the vision quality and cross-modal alignment issues existing in the current audio-to-image datasets, we propose a new audio-vision-text tri-modal dataset named A2I-Set with a sophisticated pipeline. And a hand-labelled test set A2I-eval. Furthermore, we proposed a new A2I baseline AudioCanvas with FAT-Fusion. Experimental results show that despite being trained on only 323K data, AudioCanvas can achieve high-quality image generation while maintaining a high degree of image-audio alignment, even surpassing models trained on an order of magnitude more training data or a stronger backbone. We hope this research can open new avenues in the challenging task of the audio-to-image generation field.
\clearpage

\bibliographystyle{ACM-Reference-Format}
\bibliography{sample-base}

\clearpage

\appendix

\setcounter{equation}{0}
\newtcolorbox{videoCaptionBox1}{
  enhanced, 
  attach boxed title to top left={xshift=5pt, yshift=-5pt}, 
  boxed title style={colback=black, colframe=black, sharp corners}, 
  coltitle=white, 
  fonttitle=\bfseries, 
  title=Prompt for Multimodal Caption, 
  colback=white, 
  colframe=black, 
  boxrule=1pt, 
  sharp corners, 
  left=10pt, right=10pt, top=20pt, bottom=10pt, 
  fontupper=\footnotesize,
}

\newtcolorbox{videoCaptionBox2}{
  enhanced, 
  attach boxed title to top left={xshift=5pt, yshift=-5pt}, 
  boxed title style={colback=black, colframe=black, sharp corners}, 
  coltitle=white, 
  fonttitle=\bfseries, 
  title=Prompt for Caption Transfer, 
  colback=white, 
  colframe=black, 
  boxrule=1pt, 
  sharp corners, 
  left=10pt, right=10pt, top=20pt, bottom=10pt, 
  fontupper=\footnotesize,
}

\newtcolorbox{videoCaptionBox3}{
  enhanced, 
  attach boxed title to top left={xshift=5pt, yshift=-5pt}, 
  boxed title style={colback=black, colframe=black, sharp corners}, 
  coltitle=white, 
  fonttitle=\bfseries, 
  title=Prompt for Image Filtering, 
  colback=white, 
  colframe=black, 
  boxrule=1pt, 
  sharp corners, 
  left=10pt, right=10pt, top=20pt, bottom=10pt, 
  fontupper=\small,
}

\newtcolorbox{humanevalBox}{
  enhanced, 
  attach boxed title to top left={xshift=5pt, yshift=-5pt}, 
  boxed title style={colback=black, colframe=black, sharp corners}, 
  coltitle=white, 
  fonttitle=\bfseries, 
  title=Human Evaluation Guideline, 
  colback=white, 
  colframe=black, 
  boxrule=1pt, 
  sharp corners, 
  left=10pt, right=10pt, top=20pt, bottom=10pt, 
  fontupper=\small,
}

\begin{table*}[htbp]
  \centering
  \caption{Audio Event Classes in A2I-Set}
  \label{tab:class_list}
  \scriptsize       
  \setlength{\tabcolsep}{2.5pt} 
  \begin{tabular*}{\linewidth}{@{\extracolsep{\fill}} llllllll}
    \toprule
    \multicolumn{8}{c}{\textbf{241 Audio Event Classes}} \\
    \midrule
    DJ & accordion & air conditioning unit & aircraft &
    aircraft engine & airplane & alarm & ambulance \\
    animal & artillery & artist & audience &
    baby & background music & backing vocalist & badminton \\
    bagpipe & ball & band & band instrument &
    banjo & basketball & bass & bass drum \\
    bell & bicycle & bird & blender &
    boat & bongo drum & bowl & brass instrument \\
    breath & bus & camera & car &
    cat & cello & chainsaw & chicken \\
    child & chime & choir & chorus &
    church bell & clap & clarinet & click \\
    clock & computer & conga drum & container &
    conversation & cow & cowbell & crow \\
    crowd & cutlery & cymbal & dancer &
    didgeridoo & digital drum & dish & dog \\
    door & drawer & drill & drum &
    drum machine & drum set & drumming & duck \\
    electric bass & electric device & electric guitar & electronic drum &
    electronic music & emergency vehicle & engine & explosion \\
    fabric & fan & faucet & female &
    female speaker & female vocalist & fighter jet & fire \\
    fire truck & firework & flute & fog horn &
    food & footstep & french horn & frog \\
    glass & glockenspiel & goose & guitar &
    gun & hair dryer & hammer & hand \\
    harmonica & harmonium & harp & harpsichord &
    heavy vehicle & helicopter & horn & horse \\
    impact noise & impact sound & insect & keyboard &
    kick drum & kitchenware & knife & laughter \\
    lawn mower & leaf & lion & liquid &
    machine & male & male speaker & male vocal \\
    male vocalist & male voice & mallet percussion & mandolin &
    marching band & marimba & mechanical equipment & metal object \\
    microphone & motor vehicle & motorboat & motorcycle &
    mouse click & music & musical Instrument & musical instrument \\
    musician & object & orchestra & organ &
    paddle & pan & paper & percussion instrument \\
    performer & person & piano & pig &
    pigeon & plastic & player & police car \\
    power tool & printer & racing car & radio &
    rain & rapper & river & rocket \\
    roller coaster & runner & saxophone & sewing machine &
    shaker & sheep & shoe & singing bowl \\
    siren & sitar & skateboard & snowmobile &
    soloist & speaker & speaking person & speech \\
    spray & spray can & steam train & steel drum &
    steelpan & string & string instrument & stringed instrument \\
    subway & synth instrument & synth pad & synth string &
    synthesizer & tabla & table tennis & tambourine \\
    tap & tap dancer & telephone & tennis ball &
    thunder & tick & timpani & tire \\
    toilet & tool & toy & tractor &
    traffic & train & train horn & truck \\
    trumpet & tuba & turntable & typewriter &
    ukulele & utensil & vacuum cleaner & vehicle \\
    video game & violin & vocal & vocal ensemble &
    vocalist & voice & water & watercraft \\
    waterfall & wave & whistle & wind &
    wind chime & wood & woodwind instrument & xylophone \\
    zipper & & & \\
    \bottomrule
    \label{tab:classes}
  \end{tabular*}
\end{table*}

\section{Additional Details of A2I-Set}
\begin{figure*}[htbp]  
    \centering  
    \includegraphics[width=\textwidth]{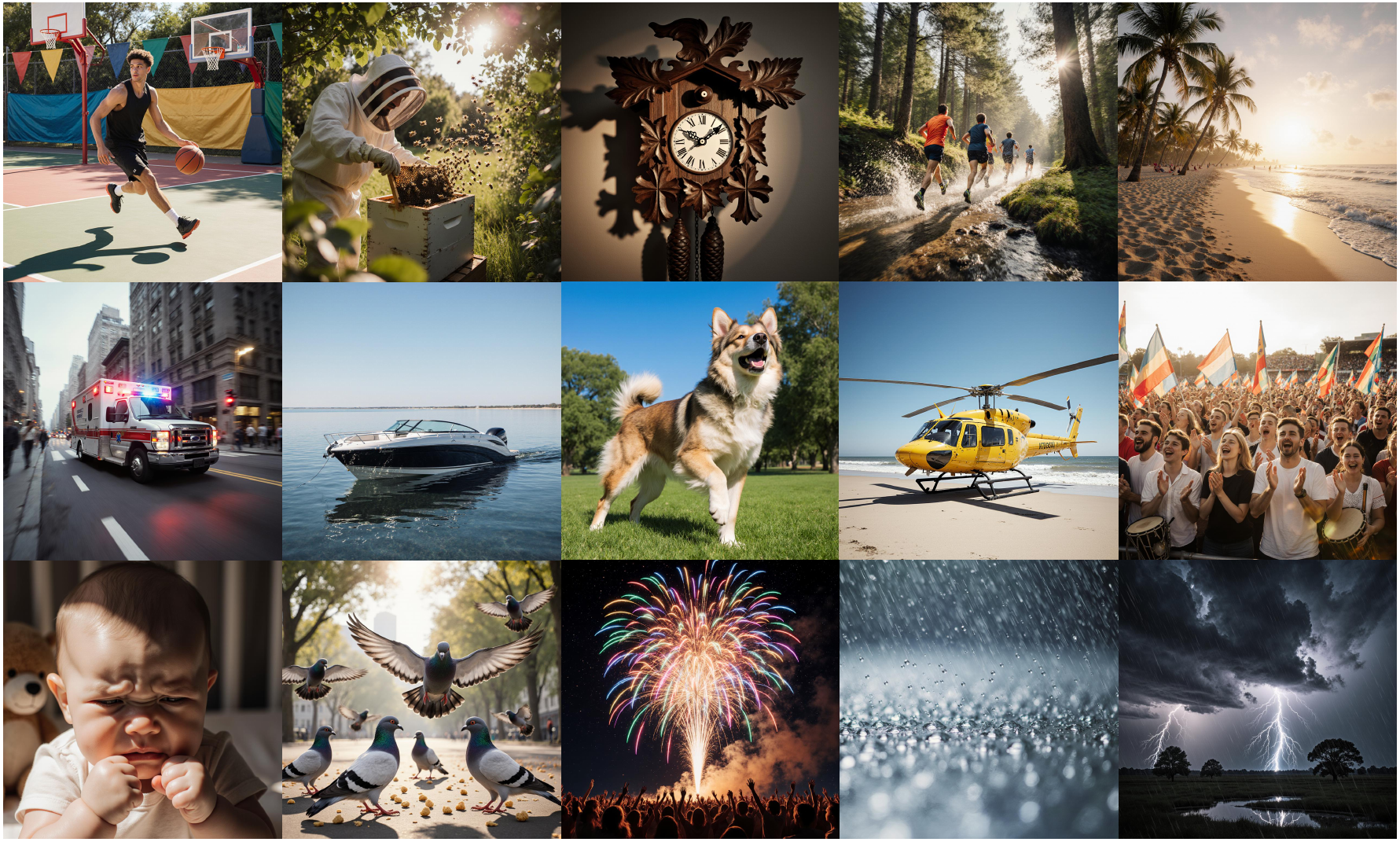}
    \includegraphics[width=\textwidth]{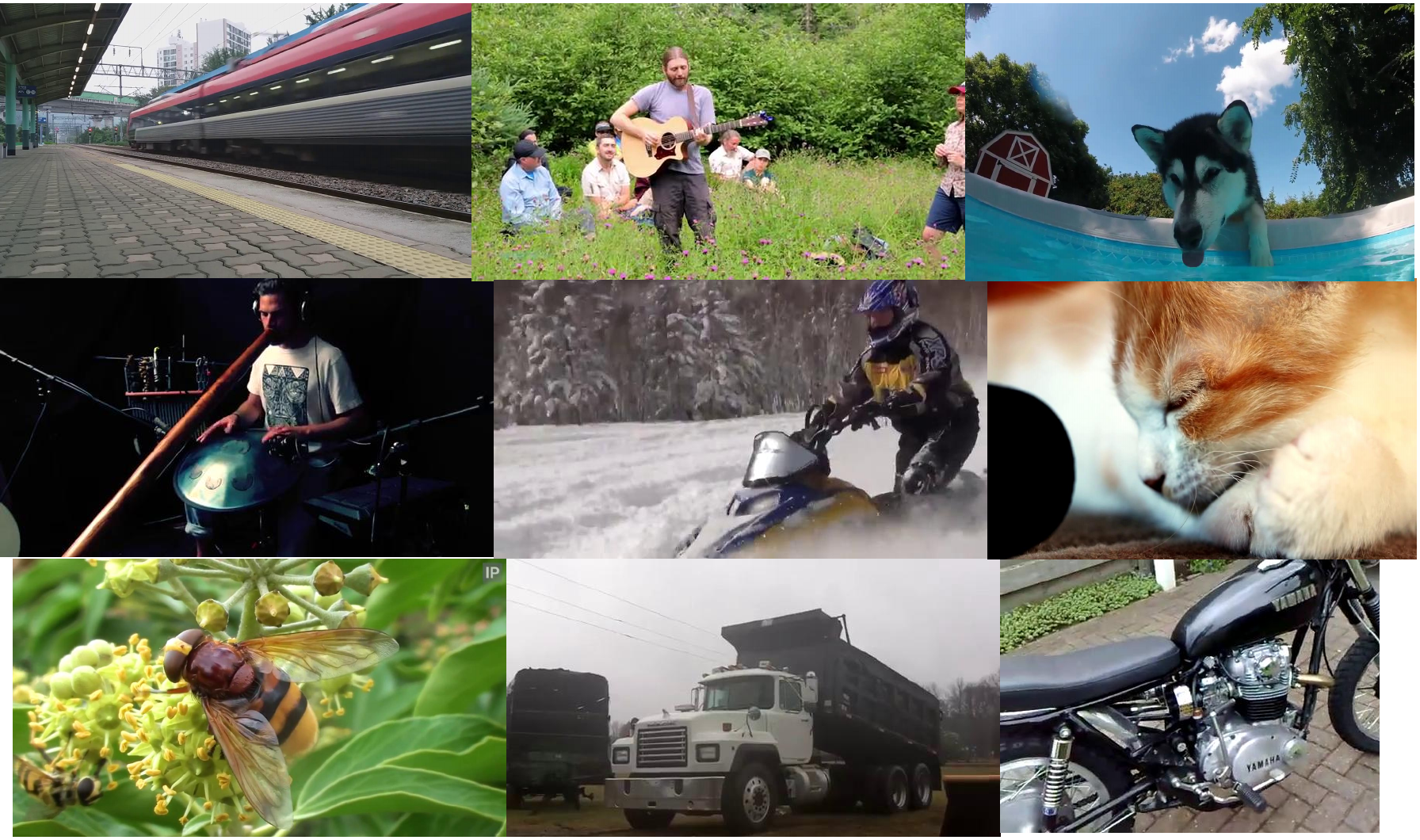}
    \caption{Additional image examples of A2I-Set.}  
    \label{fig:addition data}  
\end{figure*}

In this section, we further elaborate extra details of A2I-Set as well as the implementation process in the subsequent subsections. Finally, in \cref{fig:addition data}, we present some additional examples of A2I-Set, which further prove the high quality of our dataset.
\label{sec:dataset construction}

\begin{figure}[tbp]  
    \centering  
    \includegraphics[width=\linewidth]{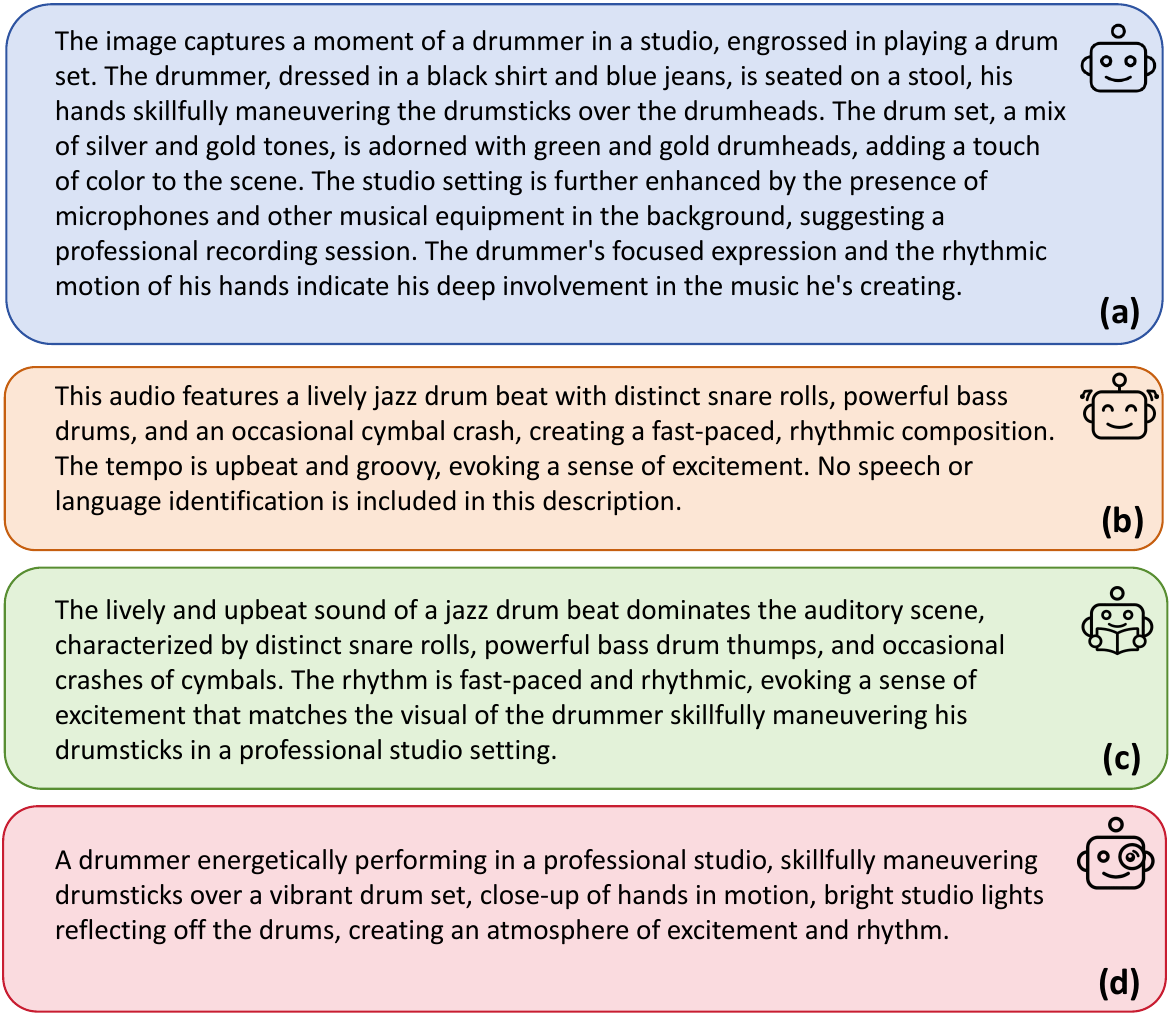}
    \caption{A complete example of caption transformation. In the Multimodal Caption stage, video caption (a) and audio caption (b) are firstly transformed into an audio caption with video message (c). To fit Text2image model input, it is then utilized to construct audio related image instruction (d).}  
    \label{fig:caption}  
\end{figure}

\label{sec:dataset analysis}
\subsection{Audio Classes Distribution}
\begin{figure}[tbp]  
    \centering  
    \includegraphics[width=\linewidth]{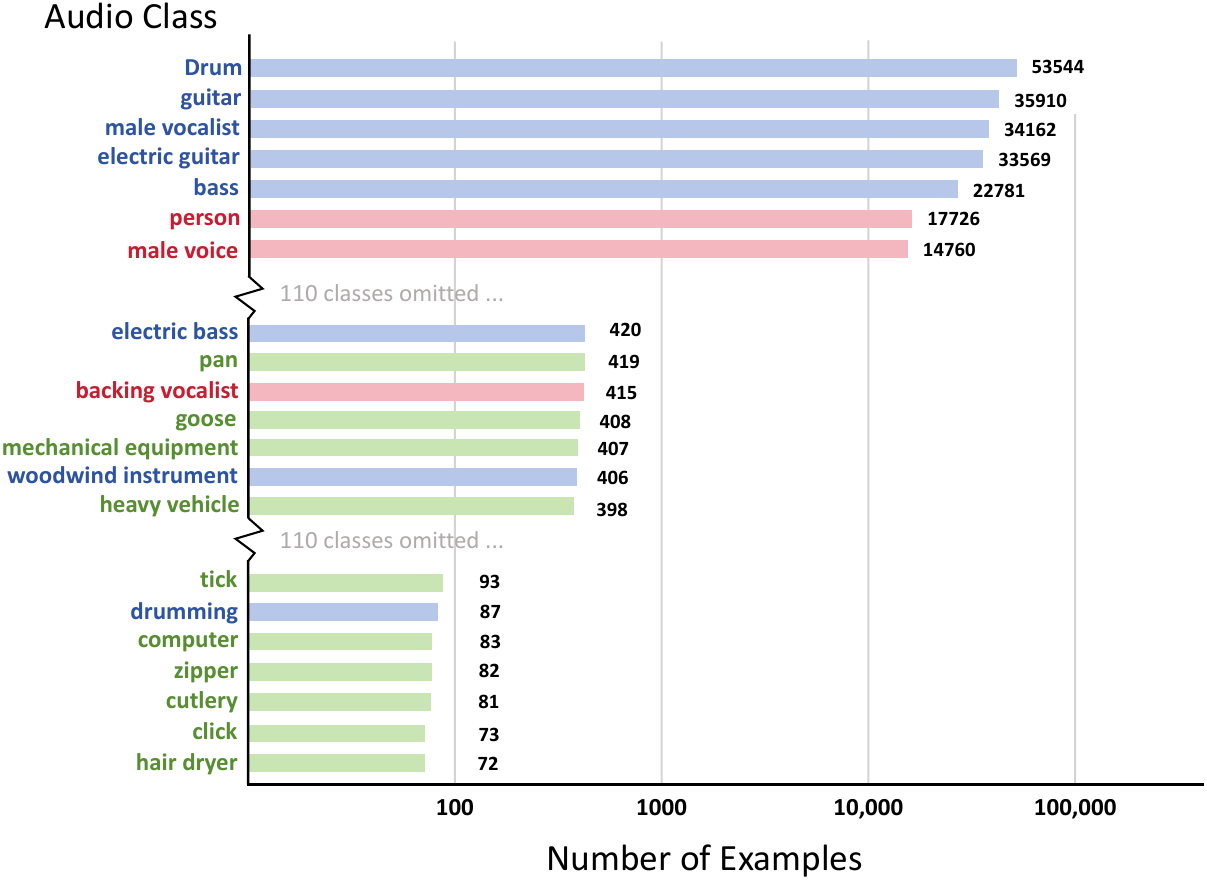}
    \caption{The correspondence between audio categories and image counts in A2I-240K-train. Note that one piece of data can include multiple audio categories. Some categories are omitted to save space. Music, speeches, and sounds are displayed in different colors.}  
    \label{fig:distri}  
\end{figure}

In \cref{tab:classes}, we display the 241 audio event classes that appear in A2I-Set, noting that there may be concept overlap within some categories due to the hierarchical characteristics in A2I-Set. In \cref{fig:distri}, we show the detailed correspondence between audio categories and total image counts in A2I-Set. We observe that the number of data points for different audio categories is unbalanced. This occurs because our data sources AudioSet and VGGSound exhibit similar characteristics. For this reason, we further present A2I-balanced to encourage research that requires balanced data. A2I-balanced includes approximately 50 images per class selected from the synthetic images in A2I-Set, with a total of 10K images. In later chapters, we use the A2I-balanced subset to further demonstrate the broad applicability of A2I-Set.

\subsection{Multimodal Caption}
Before performing the caption task, we find during sampling inspection that the quality of a few audio files is poor, which will affect the captioning effect of the audio model. Therefore, we first performed audio filtering before the Multimodal Caption. We have set three filtering dimensions: 1. Duration filtering: the audio length must be at least 4.0 seconds to eliminate overly short segments lacking effective content. 2. Silence filtering: we calculate the audio's Root Mean Square (RMS) energy frame by frame, and mark frames with RMS $<$ 0.01 as "silent frames", finally we filter out data where the proportion of silent frames of the total duration exceeds 35\%. 3. Signal-to-Noise Ratio (SNR) filtering: we use short-time RMS energy to distinguish signal from noise, and adopt 1.5 $\times$ median energy as the separation threshold between signal and noise to ensure audio clarity and eliminate samples with excessive noise.

Given audio-video pairs, we use \textbf{Qwen2-Audio-7B-Instruct} to caption the audio. In this step, in order to fully utilize the model's ability, we simply prompt the model to describe the content within the audio. For video data, we use \textbf{PLLaVA-34B} to caption the video data, with all settings kept as the default setting of PLLaVA. After obtaining the captions of the two modalities, we perform caption fusion using \textbf{GPT-4o-mini} to get an audio caption with the potential information of the video. The input instruction can be found in \cref{fig:cap1} and the example of caption process can be found in \cref{fig:caption}. 

FusionAudio-1.2M \cite{chen2025fusionaudio} creates fine-grained audio captions through a novel multimodal context fusion pipeline. We find that the audio captions in FusionAudio-1.2M also carry implicit visual information, which may be due to visual caption fusion. So we use it as a supplement to our caption after the Multimodal Caption Stage. Due to the inability to obtain some audio sources, we finally obtain approximately 1.1M of valid audio data to be processed.

Then, we adopt CLAP \cite{laionclap2023} for audio-text pairwise data filtering. Specifically, we use CLAP's audio encoder and text encoder to process the audio and the obtained audio captions respectively, and calculate the cosine similarity between the two. For FusionAudio-1.2M caption derived from AudioSet, we choose to filter data pairs with a similarity less than 0.4, while for the data we constructed from VGGSound, we selected to filter those with a similarity less than 0.3. As we want to retain a large amount of natural sound data in VGGSound. Finally, we filtered out nearly 40\% of the total data.

\subsection{Synthetic Image Generation}
In practice, we find that directly feeding audio captions with video messages into a text-to-image model, specifically FLUX.1-Krea-dev \cite{flux1kreadev2025}, does not yield good results. We believe this is because text-to-image models sometimes struggle to understand abstract and complex audio captions. Therefore, before inputting the caption, we use \textbf{GPT-4o-mini} to transform it and incorporate a large amount of visual element information, ultimately obtaining the audio-related image instruction. Specifically, we used the prompt in \cref{fig:cap2} and constructed a high-quality image generation instruction by prompting the model to perform sounding objects identification, visual description generation, and atmosphere recognition (including lighting, color, composition, etc.).

We also present the instructions for visual large models in the VLLM Cross-Check stage in \cref{fig:cap3}. In practice, we find that directly prompting the model to identify whether an image is a real photograph yields better results than asking the model to identify unreasonable parts in the image. The model has quite an accurate recognition ability for images with non-realistic stylism, and also has good ability to distinguish unreasonable object relationships. However, even the most powerful open-source VLMs currently available struggle to perfectly identify defects in images. We found that the models perform poorly in recognizing facial distortions and the number of limbs. Although such images are in the minority in the sampling evaluation, we must still acknowledge that the images in A2I-Set cannot guarantee 100\% perfection.

\subsection{Real Image Extraction}
In the Real Image Extraction stage, we restrict the resolution to above 540$\times$360 and set the video’s minimum frame rate to 12 fps. Besides, we restrict the aspect ratio of source videos to stay between 1:3 and 3:1. This is because images with extremely skewed ratios are difficult to use in training. For jitter detection, we calculate dense optical flow within each scene on a per-shot basis and use the median value to reduce the influence of extreme values. To further refine image quality, we filter out images with an ARNIQA score below 0.4, and finally obtain approximately 150k real image candidates.

\subsection{Audio-Image Alignment Filter}

To ensure audio-visual alignment of the dataset, we introduce Sam Audio \cite{shi2025sam} to perform object-level separation on the audio. Specifically, we take the previously annotated sound-emitting objects as prompts and input them individually into sam-audio-base. We perform root mean square (RMS) energy filtering on the separated results to ensure that the target element is present in the audio. Next, to determine whether the target elements exist in the image, we use Qwen3-VL-Flash \cite{bai2025qwen3} and Ground DINO \cite{liu2023grounding} respectively to locate each target element. For Ground DINO, we set BOX\_THRESHOLD and TEXT\_THRESHOLD to 0.3 to perform the grounding task. For Qwen3-VL-Flash, we simply prompt it to perform the grounding task and output None for objects that are not found.

\begin{figure*}[p] 
\centering 
\begin{videoCaptionBox1}
\textbf{Core Task}

Given a [VIDEO\_CAPTION] generated by a visual language model and an [AUDIO\_CAPTION] generated by an audio language model. 

Your goals are:

Fuse: Integrate the information from the two captions.

Refine: Utilize the video caption to enhance, correct, and specify the audio caption, eliminating any potential ambiguity or hallucination in the audio description.

Output: Generate a pure final audio caption that only describes auditory perceptions.

\textbf{Processing Steps}

Please strictly follow the following thought chain for processing and complete these steps in your mind, and finally output only the result in the specified format:

Analysis and Preliminary Judgment:

Step 1.1: Analyze [AUDIO\_CAPTION], identify the core sound events described therein (such as speaking, music, impact sounds), sound characteristics (such as fast rhythm, high pitch), and the environment (such as sounds like it's indoors).

Step 1.2: Analyze [VIDEO\_CAPTION], identify the key visual elements therein (such as people, vehicles, animals), actions (such as running, playing musical instruments), and the environment (such as streets, forests, concert halls).

Cross - Validation and Information Fusion:

Step 2.1 - Confirmation and Enhancement: Look for parts where the video and audio information can corroborate each other. If the audio description is "the sound of a vehicle driving" and the video description is "a red sports car accelerating on the track", then the final description can be enhanced to "the roaring engine sound of a sports car accelerating".

Step 2.2 - Ambiguity Elimination: When the [AUDIO\_CAPTION] description is ambiguous (for example, "a burst of noise"), use the [VIDEO\_CAPTION] (for example, "a plate dropped on the ground and broke") to make it specific to "the sound of a plate breaking".

Step 2.3 - Conflict Resolution and Hallucination Correction:

If the [AUDIO\_CAPTION] describes an event (such as "the barking of a dog"), but there is relative but conflict visual entity in the [VIDEO\_CAPTION] (the video shows a cat), this is likely a hallucination of the audio model. At this time, the visual information should be given priority, and the description of sound should be ignored.

Conversely, if a clear sound is heard (such as "the siren sound"), but there is no possible conflict visual entity mentioned in the video caption, the sound is still an objective fact (it may be off-screen). At this time, it should be described as "the siren sound is heard in the background". 

Constructing the Final Audio Caption:

Step 3.1 - Fact Synthesis: Based on the above analysis, integrate all the verified and corrected auditory facts.

Step 3.2 - Language Style:

Objective and Accurate: Only describe what can be heard. Do not over-extend or make unfounded guesses based on the input information.

Rich in Details: Incorporate the sound details inferred from the visual information (such as the sound source, the dynamic changes of the sound).

Cautious in Wording: Try to avoid information that cannot be completely determined. If necessary, use conservative expressions such as "it sounds like...", "it may be accompanied by...", "it implies...".

If the input information is extremely scarce or severely conflicting, making it impossible to construct a reliable auditory fact, please directly output a specific string: [UNCERTAIN].

\textbf{Example}

Input 1:

[VIDEO\_CAPTION]: "A man is sitting in front of a wooden table with a laptop in front of him. He is quickly typing on the keyboard and occasionally picks up the cup next to him to drink water."
[AUDIO\_CAPTION]: "Continuous tapping sounds and occasional liquid flowing sounds can be heard, and there seems to be a slight wind sound in the background." 

\textbf{Example Output 1 Format}:

"final\_caption": "The continuous and rapid typing sounds of the keyboard form the main auditory scene, interspersed with the swallowing sound and the slight sound of liquid sloshing when drinking water."

Input 2:

[VIDEO\_CAPTION]: "The camera pans slowly, revealing a peaceful lakeside park with shady trees and several ducks swimming on the lake. There are no man-made vehicles in the frame except for the natural scenery."
[AUDIO\_CAPTION]: "Clear bird chirping and the rustling sound of leaves in the gentle breeze can be heard. In the distance of the background, there seems to be the rumbling sound and whistle of a passing train."

\textbf{Example Output 2 Format}:

"final\_caption": "A peaceful natural soundscape, with the foreground featuring the clear and melodious chirping of birds and the gentle rustling of leaves in the wind. In the background of these sounds, the rumbling of a train and a long whistle can be heard coming from a distance, suggesting that the world outside the park is still in motion."

Now please process the following input:

\end{videoCaptionBox1}
\caption{Prompt for Multimodal Caption.}
\label{fig:cap1}
\end{figure*}

\begin{figure*}[p] 
\centering 
\begin{videoCaptionBox2}
\textbf{Core Task}

For a text describing an audio scene \textbf{[Audio Scene Description]}, parse out the sound-producing subject within it, and with this subject as the center, generate a visual generation instruction that describes details such as the appearance, actions, and environment of this subject, in order to provide inspiration for an AI painting model. Please think strictly in accordance with the following steps:

1. \textbf{Subject Identification}: Identify the main sound-producing subject (such as a person, an animal, an object, etc.) from the text. Sometimes the text may contain some visual elements. In this case, consider both visual and auditory elements comprehensively. For elements with uncertainties in the description, carefully consider whether to include them in combination with other definite elements. If the descriptions of visual and auditory elements are related but different, give priority to visual elements. If there is no practical connection between the two, you can output "$<$UNCERTAIN$>$" and end the output directly.
After successfully identifying the subject, output the name of the subject in the following format: $<$Subject Name$>$; and output.

2. \textbf{Visual Description}: Based on the previously identified subject, generate a concise and reasonable visual description, including the appearance characteristics, actions, and environment of the subject. While meeting the visual expression needs, avoid adding unnecessary or unhinted visual details. If the relevance between the subjects is poor, choose a more representative subject for description. At this time, the secondary subjects can be ignored in the subsequent process.

3. \textbf{Atmosphere Identification}: If there is a description of the auditory atmosphere, transform it into a visual atmosphere (lighting, color, composition). Grasp the "emotional tone and scene characteristics" in the auditory sense, and then match them through visual elements such as the intensity of light and shadow, the warmth and coolness of colors, and the density of composition, so that the two senses resonate.

\textbf{Note}: The final generated visual generation instruction needs to completely eliminate any words that directly describe sounds. At the same time, the final described scene must be a realistic one, and any description of abstract or non-realistic symbol is not allowed.

\textbf{Processing Steps}

Please strictly follow the following thought process to convert the auditory description into a visual instruction:

Identify Core Elements: First, break down the input \textbf{[Audio Scene Description]} and identify:

Sound Emitters: What is in fact making the sound according to the description? Selectively convert the certain sound emitter into a specific visual object. (e.g., "keyboard sound" → $<$keyboard$>$. "footsteps" → $<$feet and the ground$>$. "guitar melody" → $<$guitar$>$) Combine \textbf{all visual elements} and judge whether to adopt all or part of them to form a reasonable and concise scenario.

Acoustic Actions: What action is the sound emitter doing? (e.g., $<$keyboard$>$ → rapid tapping, $<$vehicle$>$ → accelerating roar, $<$piano$>$ → gentle strumming, $<$machine$>$ → running)

Auditory Environment \& Atmosphere (selectively do): What is the sense of space and the feeling the sound gives? (e.g., open and reverberant, noisy and chaotic, peaceful and serene, tense and mysterious)

Conduct Visual Translation:

Atmosphere → Style/Lighting (selectively do): Translate the auditory atmosphere into visual language. For example,

"Open and reverberant" → "A vast space, minimalist composition, cool colors, and perhaps long shadows".

"Noisy and chaotic" → "Crowded composition, overlapping elements, high contrast, and perhaps motion blur".

"Peaceful and serene" → "Soft light, symmetrical composition, warm or natural colors, and a clean picture".

Action → Dynamics/Composition (selectively do): Convert the acoustic action into motion or action in the description. Additionally, determine composition and viewpoint using simple composition rules (centered, diagonal, or rule-of-thirds), and ensure the sound source remains the visual focus with minimal distractions. For example,

"Rapid tapping" → "Close-up of hands with motion blur at the fingertips".

"Accelerating roar" → "Vehicle with speed lines and a blurred background, wide angle view".

"Gentle strumming" → "Close-up of fingers on guitar strings, soft light, and a focused atmosphere".

Use AI Painting Grammar: Adopt the grammar preferred by AI painting models, usually keywords or phrases separated by commas. Note that the text length should not exceed 77 tokens. The visual generation instruction should be concise and clear, avoiding unnecessary words.

\textbf{Example}

Input 1:

[Audio Scene Description]: "The continuous and rapid typing sounds of the keyboard form the main auditory scene, interspersed with the swallowing sounds when drinking water and the slight sounds of liquid sloshing, implies a scene where an employee is working overtime."

\textbf{Example Output 1 Format}:
{{
"[Main Subject]": "$<$keyboard$>$, $<$liquid$>$, $<$swallowing$>$",
"[Visual Generation Instruction]": "A person's hands typing rapidly on a mechanical keyboard, close-up shot, a glass of water sits on the desk beside the keyboard, shallow depth of field, focused on the fingers."
}}

Input 2:

[Audio Scene Description]: "A solo melody played by an acoustic guitar, close-up shot, with a clear and bright timbre. The performer creates soothing and textured phrases through strumming and fingerpicking techniques, the environment is very quiet and the overall musical atmosphere is serene and focused."

\textbf{Example Output 2 Format}:
{{
"[Main Subject]": "$<$guitar$>$",
"[Visual Generation Instruction]": "A musician sitting on a stool on a dimly lit stage and playing guitar, a single warm spotlight illuminates him, intimate and focused atmosphere, high contrast, dark background."
}}

Now please process the following input:

\end{videoCaptionBox2}
\caption{Prompt for Caption Transfer.}
\label{fig:cap2}
\end{figure*}

\begin{figure*}[ht] 
\centering 
\begin{videoCaptionBox3}

Play the role of a professional image analyst and strictly evaluate the authenticity of the following picture. Some of the pictures are fake images or digital artworks generated by AI model while other pictures are real-world photos taken by a camera. You need to evaluate and identify them carefully.
Your evaluation should be based on the general public's perception of the real world. You should consider multiple aspects, if there is an abnormality in the picture, the picture is considered fake:

Here are some reference points for your evaluation, but you are not limited to these points, try to be as comprehensive as possible:

Logical and Contextual Sanity:

Scene Plausibility: Is the central scene in the image logical? For example, are birds appearing in inappropriate indoor environments (like an office or bedroom)? Are large mammals in illogically confined spaces?

Object Relations: Are the relationships between objects in the image reasonable? For instance, are the size proportions of objects consistent? Do their positions and interactions make sense?

Physical World Consistency:

Light and Reflection: Is the direction of the light source consistent? Do the direction, length, and sharpness of shadows match the light source? Are the reflections and refractions on object surfaces consistent with their materials and the environment?

Physical Dynamics: If the image contains dynamic elements (e.g., flowing water, fire, moving objects), do their forms and trajectories adhere to the laws of physics?

Detail and Structural Integrity:

Biological Features:

Humans: Are facial features (eyes, teeth, ears) symmetrical and natural? Is the number, length, and shape of fingers and toes correct? Are the connections and bending of limbs anatomically correct? Is the skin texture overly smooth or exhibiting anomalies? Is the face clearly visible without distortion or blurring?

Animals: Is the animal's body structure complete and correctly proportioned? Is the texture of fur or feathers natural?

Object Structure: Do objects appear strange due to unnatural distortion, stretching, or compression? Are the edges and joints of objects clear and logical?

Absence of Anomalous Visual Elements:

No Abstract Symbols or Patterns: Does the image contain geometric patterns, symbols, text, or abstract visual effects that do not exist in the real world and have no clear meaning? Does the background or object surfaces contain unexplainable, repetitive, or strange textures?

Color Realism: Is the overall tone and color saturation of the image natural? Are there illogical, overly vibrant, or bizarre color combinations? Is the color tone of the picture a common one in the real world?

More: Anything else that seems out of place or inconsistent with the real world.

Finally, consider the question above and directly output "Real" when you definitely believe the picture is from real world or "Fake" for else situations, outputting anything else.

\end{videoCaptionBox3}
\caption{Prompt for Image Filtering.}
\label{fig:cap3}
\end{figure*}

\section{Details of Selected Metrics}
To comprehensively measure the visual quality of generated images, we use typical quantitative metrics, including the Fréchet Inception Distance (FID) and the Inception Score (IS). Besides, evaluating the vision quality of image generation models requires alignment with human perception. Thus, we introduce two human preference metrics, Aesthetic evaluation score (Aes) and Human Preference Score v3 (HPSv3). For cross-modal alignment evaluation metrics in this work, we use Audio-Image Similarity (AIS) and Text-Image Similarity (TIS). Both metrics aim to measure the similarities of the paired data by projecting them to the shared embedding space.

\textbf{Fréchet Inception Distance (FID)} measures the distribution discrepancy between the real images $X_r$ and the generated images $X_g$ in the feature space of an Inception-V3 network. Let $f(\cdot)$ denote the Inception-V3 \texttt{pool3} feature extractor. For the real and generated features, we compute the mean of image features $\mu_r$ and $\mu_g$ and the covariance of image features $\Sigma_r$ and $\Sigma_g$:

\begin{equation}
\begin{split}
\mu_r = \mathbb{E}[f(X_r)], \quad \Sigma_r = \operatorname{Cov}(f(X_r))\\
\mu_g = \mathbb{E}[f(X_g)],  \quad \Sigma_g = \operatorname{Cov}(f(X_g)).
\end{split}
\label{eq:6}
\end{equation}
FID is then defined as the Fréchet distance between the two Gaussians distribution:

\begin{equation}
\operatorname{FID}(X_r, X_g) = \|\mu_r - \mu_g\|_2^2 + \operatorname{Tr}\left( \Sigma_r + \Sigma_g - 2(\Sigma_r \Sigma_g)^{1/2} \right).
\label{eq:7}
\end{equation}

where $\operatorname{Tr}$ denotes the trace of a matrix, and $\|\cdot\|_2^2$ denotes the Euclidean norm.
Lower FID indicates that the generated images are closer to the real data distribution in both feature statistics and visual diversity. In our work, we uniformly use precomputed distribution of ImageNet to calculate FID.

\textbf{Inception Score (IS)} evaluates both the image quality and the class diversity of the generated images using the \textit{Inception-V3 classifier}. IS converts the predicted class distribution $p(y|X_g)$ of generated images $X_g$ into quantifiable scores through the pre-trained Inception-v3 image classification model. The marginal class distribution can be denoted as:

\begin{equation}
p(y) = \sum_i p(y|x_i) \quad where \quad x_i \in X_g
\label{eq:8}
\end{equation}
IS is defined as:

\begin{equation}
\text{IS} = \exp\left( \mathbb{E}_{x \sim p_g} \left[ D_{\text{KL}} \left( p(y|x) \parallel p(y) \right) \right] \right)
\label{eq:9}
\end{equation}

where $D_{\text{KL}}$ denotes the KL divergence, which measures the degree of difference between two probability distributions. IS is a measure of the clarity and diversity of generated images, and a larger IS value is better. IS depends on the ImageNet classifier and may not fully reflect semantics of non-ImageNet domains. Thus IS is often used as an auxiliary metric alongside FID.

\textbf{Aesthetic evaluation score (Aes)} represents the aesthetic score of generated images. Specifically, we use Aesthetic Predictor V2 \footnote{https://github.com/christophschuhmann/improved-aesthetic-predictor} to predict a score range in 1-10 with a higher score representing a more attractive visual effect. Aesthetic Predictor V2 is trained on multiple human annotated aesthetic rating datasets and demonstrates strong consistency with human evaluation results. Generally speaking, the Aes score distribution of ordinary images is concentrated at 4 to 5.

\textbf{Human Preference Score v3 (HPSv3)} is introduced by \cite{ma2025hpsv3}, aiming at evaluating the performance of Text-to-Image Generation Models Aligned with Human Perception. Specifically, HPSv3 introduces over 1M pairwise comparison data covering the full spectrum from low-quality generated images to high-quality real photographic images, achieving state-of-the-art human perception alignment prediction. The HPSv3 benchmark in \cite{ma2025hpsv3} shows that Stable Diffusion v2.0 has an average HPSv3 score of -0.24. In comparison, Stable Diffusion 3 achieves a score of 5.31, and state-of-the-art text-to-image models such as Flux-dev exceed 10. Noted that although fine-tuned with the older version SD 1.4, our AudioCanvas scores 4.81 on A2I-eval.

\textbf{Audio-Image Similarity (AIS)} is initially designed by \cite{yariv2023audiotoken} to evaluate the semantic alignment between input audios and generated images through measuring their cosine similarity in a shared feature space. Specifically, Wav2CLIP \cite{wu2022wav2clip} is used to project audio inputs into shared representations. while image representations are extracted from generated images through CLIP image encoder. The higher the similarity between the two, the more the generated image fits the input audio.

\textbf{Text-Image Similarity (TIS)} has a similar idea as AIS. Modal consistency is measured by comparing the cosine similarity between the text feature corresponding to the audio and the feature of the generated images. Since A2I-Eval and Landscape contains ground truth sounding objects corresponding to the test audio, we use the CLIP text encoder to perform feature extraction on them. And we utilize the CLIP image encoder to perform feature extraction on the generated images. Similarly, the higher the similarity between the two features, the more the generated image conforms to the audio's descriptive text.

\section{A2I-Set on Other SD Models}
\begin{figure*}[htbp]  
    \centering  
    \includegraphics[width=\textwidth]{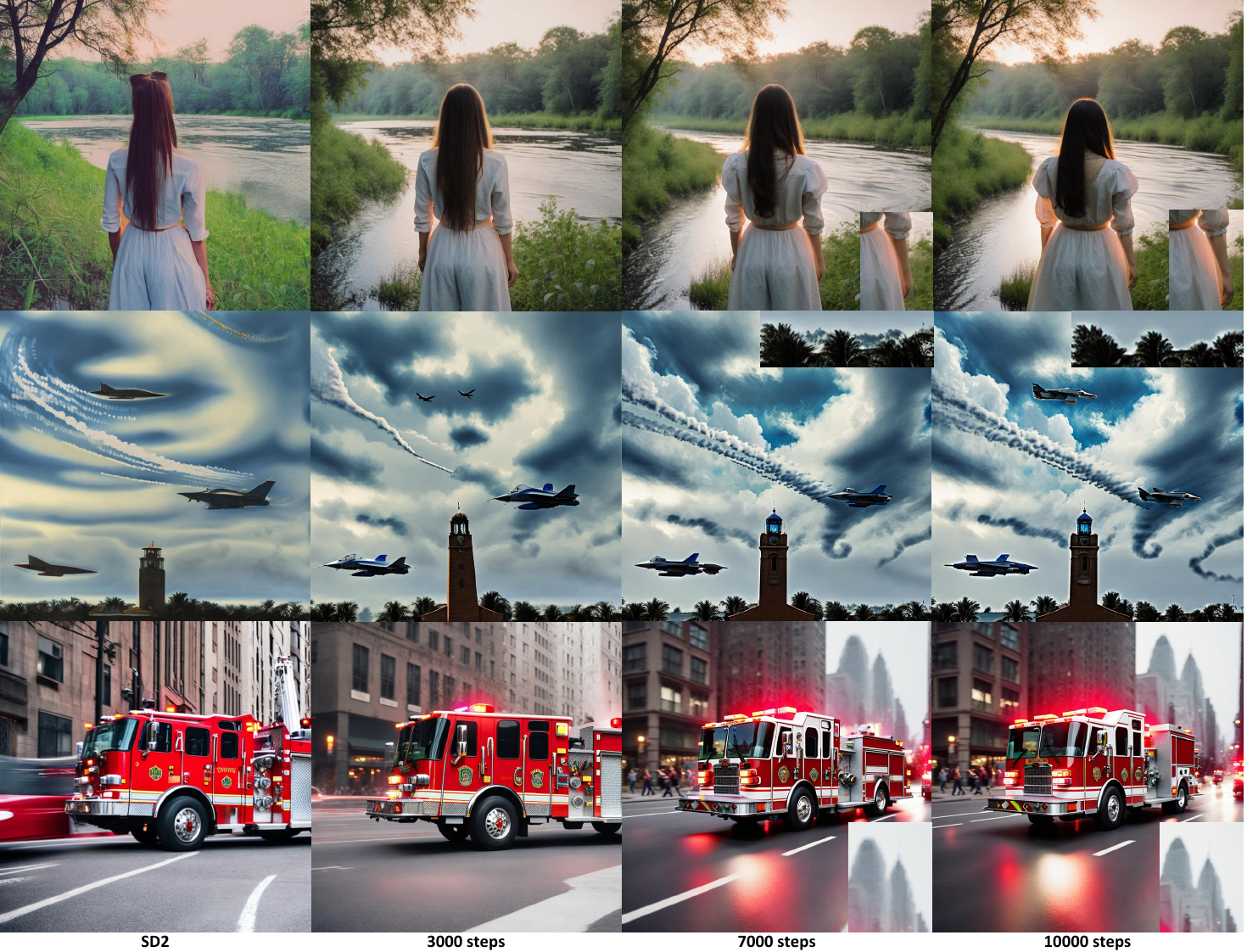}
    \caption{Test results of SD2 fine-tuned on A2I-balanced. To ensure fairness, the seeds are set to be consistent. Zoom in for a better viewing experience.}  
    \label{fig:SD2}  
\end{figure*}

In this section, we further conducted qualitative experiments on SD2 and SD-XL to demonstrate the high quality of the dataset we proposed. Specifically, considering computational resource constraints, we use image-text pairs from A2I-balanced instead of the full training set to perform full-parameter fine-tuning on both models. \cref{fig:SD2} shows the generated results of fine-tuned SD2. For SD2, we set the learning rate to 4e-7 and trained for a total of 10,000 steps on our dataset. The image resolution during training and inference is maintained at 768$\times$768. During inference, we set the denoising steps to 30 and the cfg scale to 7.5. To ensure fairness, we keep the seed consistent for the three examples across different training steps models. It can be seen that when comparing with the images generated by the original SD2, the images generated by our fine-tuned model are significantly better in generation effect. In the early stage of training, the model quickly adjusted the image style and learned the application of light and shadow knowledge. At the end of training, the model mainly focused on optimizing the details of the images. As can be seen in the last two columns of \cref{fig:SD2}, we zoom in to show some visual details in the images.

\begin{figure*}[htbp]  
    \centering  
    \includegraphics[width=0.8\textwidth]{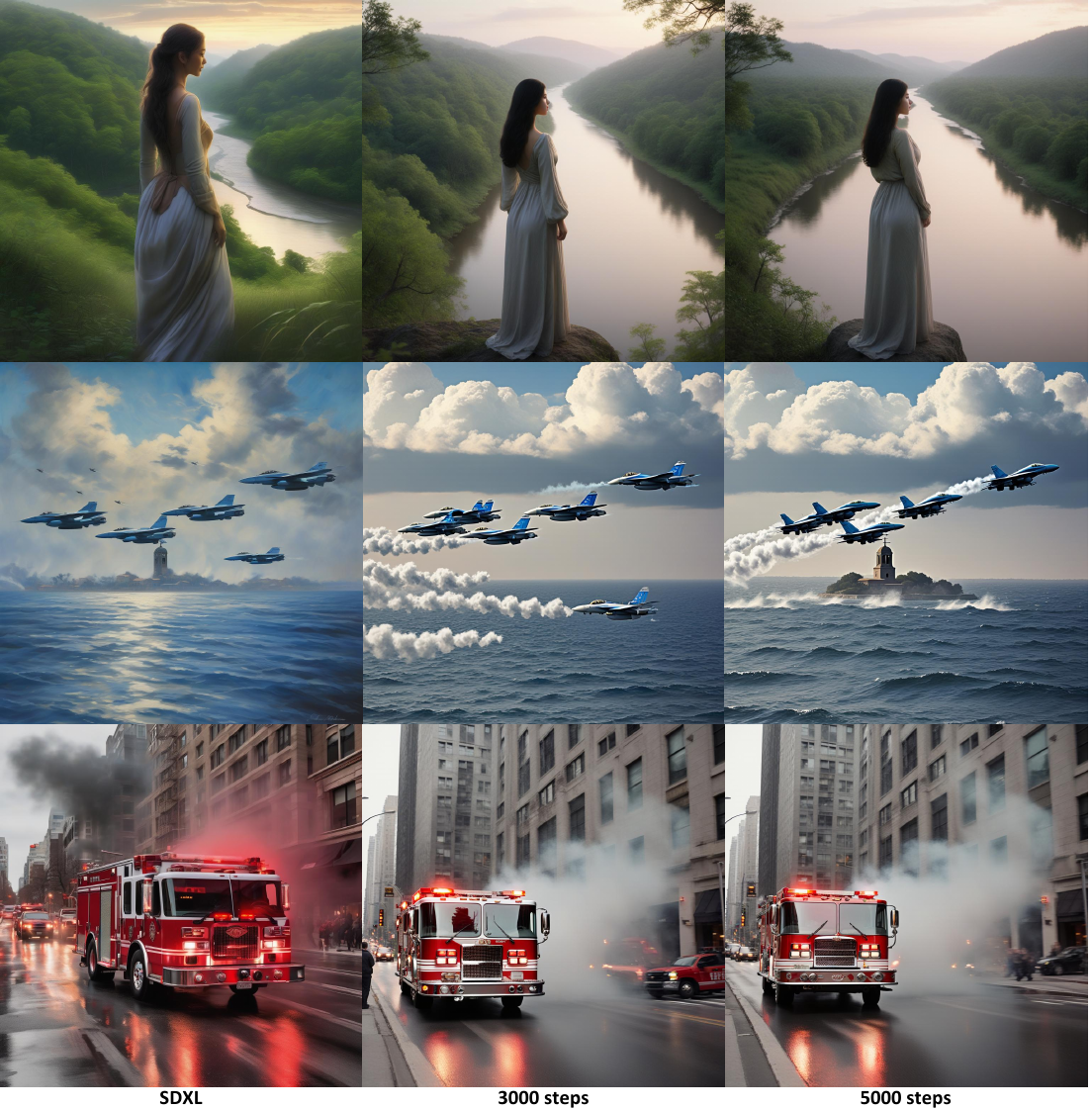}
    \caption{Test results of SD-XL fine-tuned on A2I-balanced. To ensure fairness, the seeds are set to be consistent. Zoom in for a better viewing experience.}  
    \label{fig:SDXl}  
\end{figure*}

For SD-XL, we set the learning rate to 1e-6 and trained for a total of 5,000 steps on our A2I-balanced. The image resolution during training and inference is maintained at 1024$\times$1024. However, the conclusion regarding the fine-tuning of the SD-XL model is somewhat different. As in \cref{fig:SDXl}, since the performance of SD-XL is already very powerful, the most prominent change after fine-tuning is the drawing style—from the original comic-like style to a style closer to reality. In terms of image details, we can only observe a slight improvement. This may be due to the fact that the amount of training data is too small compared to the massive training data of SD-Xl. Or it may be because our training steps are not sufficient.

\section{Preliminary on Diffusion Model}
Diffusion models are a type of deep learning model based on probabilistic generation. They gradually add Gaussian noise to data through a forward diffusion process, and then recover data from the noise through a reverse diffusion process, thereby generating new data similar to the training data. The SD v1.4 model we used is base on Latent Diffusion Models (LDMs) \cite{rombach2022high}, a class of conditional generative models that perform diffusion not directly in pixel space but in a compressed latent space learned by a variational autoencoder (VAE). Compared to pixel-space diffusion DDPM \cite{ho2020denoising}, LDMs significantly reduce computational cost while maintaining high generative fidelity.

Given an image $x_0 \in \mathbb{R}^{H \times W \times 3}$, a pretrained VAE encoder $\mathcal{E}$ maps it to a latent representation:
\begin{equation}
    z_0 = \mathcal{E}(x_0), \qquad
    z_0 \in \mathbb{R}^{h \times w \times c}.
\end{equation}
The VAE decoder $\mathcal{D}$ reconstructs the image:
\begin{equation}
    \hat{x} = \mathcal{D}(z_0).
\end{equation}
Since $h,w \ll H,W$, diffusion process is then performed on a much smaller latent tensor, which highly reduce the computational cost of DDPMs. LDMs adopt the same forward noising process as DDPMs, but applied in latent space. Noise is gradually added to $z_0$ as:
\begin{equation}
    q(z_t \mid z_{t-1})
    = \mathcal{N}\left(
        \sqrt{1-\beta_t} \, z_{t-1},\;
        \beta_t \mathbf{I}
    \right),
\end{equation}
or equivalently:
\begin{equation}
    q(z_t \mid z_0)
    = \mathcal{N}\left(
        \sqrt{\bar{\alpha}_t} \, z_0,\;
        (1 - \bar{\alpha}_t)\mathbf{I}
    \right),
\end{equation}
where $\bar{\alpha}_t = \prod_{s=1}^t (1-\beta_s)$, $\beta_t$ is the noise coefficient at step $t$, which is used to control the intensity of Gaussian noise added to the latent variable at the $t$-th step in the forward diffusion process.

In LDMs, the noise prediction model parameterized by $\theta$, specifically a denoising UNet $\epsilon_\theta$, is trained to model the reverse process:
\begin{equation}
    p_\theta(z_{t-1} \mid z_t, c)
    = \mathcal{N}\big(
        \mu_\theta(z_t, t, c),\;
        \Sigma_\theta(z_t, t, c)
    \big),
\end{equation}
where $c$ denotes a conditioning signal (e.g., text embedding).

Typically, training follows the simplified noise-prediction objective:
\begin{equation}
    \mathcal{L}_{\mathrm{simple}}
    = \mathbb{E}_{z_0,\, t,\, \epsilon}
    \Big[
        \lVert \epsilon
        - \epsilon_\theta(z_t, t, c)
        \rVert_2^2
    \Big],
\end{equation}
where
\begin{equation}
    z_t
    = \sqrt{\bar{\alpha}_t} z_0
    + \sqrt{1-\bar{\alpha}_t}\, \epsilon.
\end{equation}

Conditions such as texts, images and audios are injected through cross-attention layers inside the UNet:
\begin{equation}
    \epsilon_\theta(z_t, t, c)
    = \mathrm{UNet}_\theta(z_t, t, c).
\end{equation}

Classifier-free guidance is a guidance technique in diffusion models that does not require an external classifier. It achieves a trade-off between the quality and diversity of generated samples by jointly training conditional and unconditional diffusion models and linearly combining their score estimates during sampling. Classifier-free guidance is applied during inference:
\begin{equation}
    \epsilon_{\mathrm{guided}}
    = (1+w)\,
    \epsilon_\theta(z_t, t, c)
    - w\,
    \epsilon_\theta(z_t, t, \varnothing),
\end{equation}
where $w$ is the guidance scale.

Starting from Gaussian noise $z_T \sim \mathcal{N}(0, \mathbf{I})$, the model iteratively denoises process can be denoted as:
\begin{equation}
\begin{aligned}
    z_{t-1}
    &= \frac{1}{\sqrt{\alpha_t}}
    \left(
        z_t
        - \frac{1 - \alpha_t}{\sqrt{1 - \bar{\alpha}_t}}
        \epsilon_\theta(z_t, t, c)
    \right)
    + \sigma_t \xi,
\end{aligned}
\end{equation}
where $\xi \sim \mathcal{N}(0, \mathbf{I})$.

\begin{figure*}[htbp]  
    \centering  
    \includegraphics[width=\textwidth]{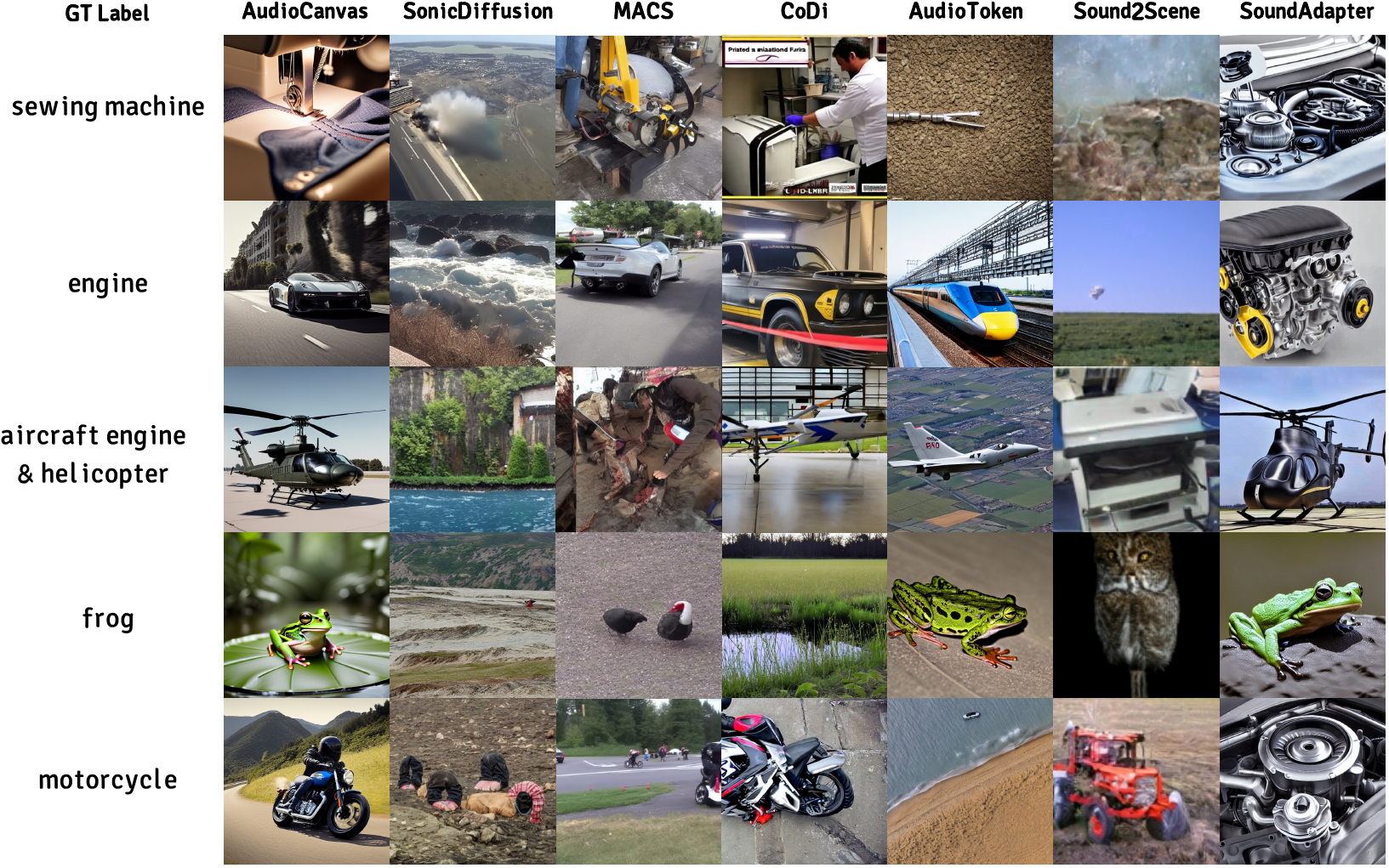}
    \vspace{5pt}          
    \hrulefill             
    \vspace{5pt}          
    \includegraphics[width=\textwidth]{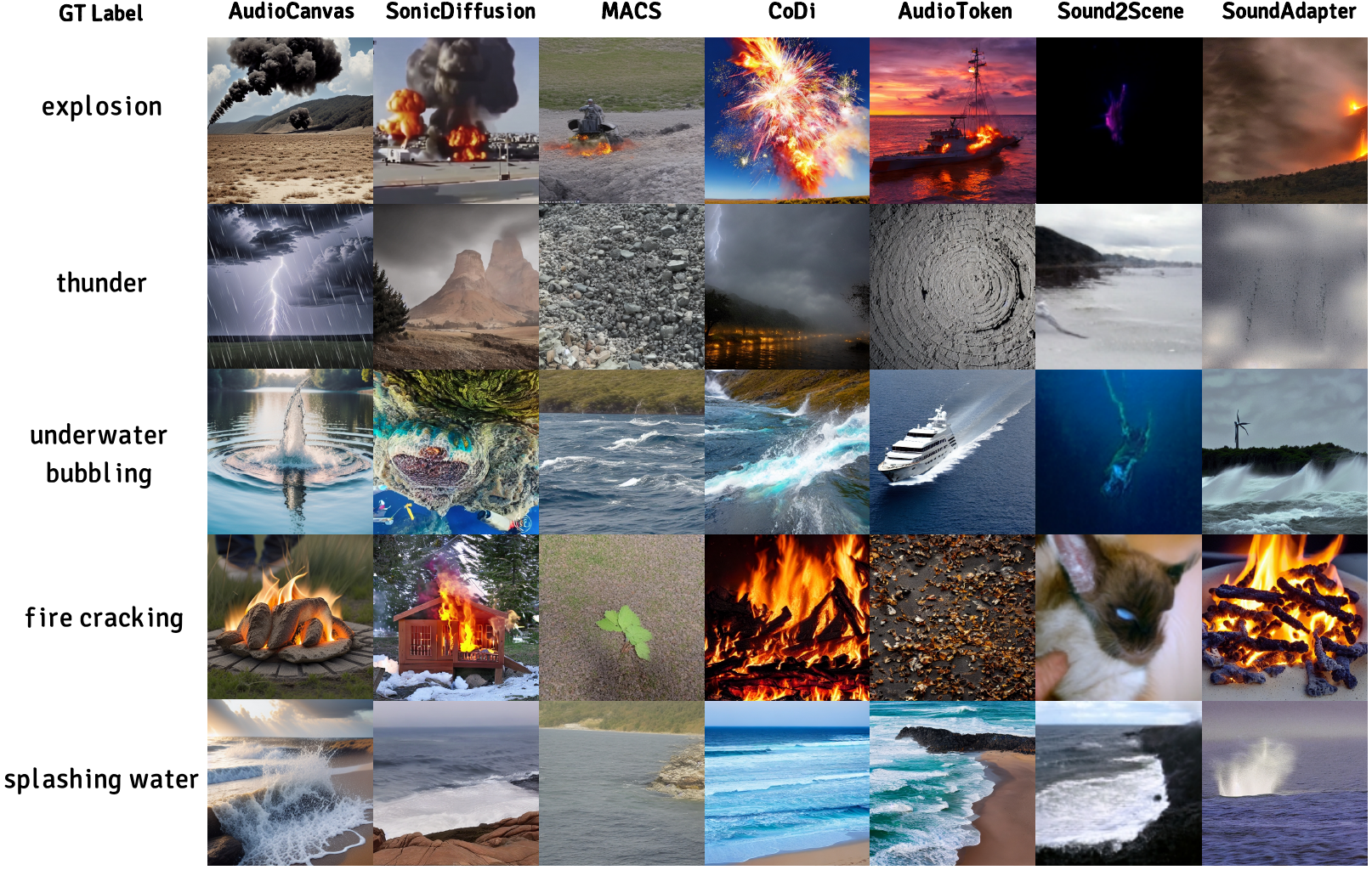}
    \caption{Comparison results of different models on A2I-eval and Landscape. Zoom in for a better experience.}  
    \label{fig:comp}  
\end{figure*}

\section{Additional Comparison with baseline}

\begin{figure*}[htbp]  
    \centering  
    \includegraphics[width=\textwidth]{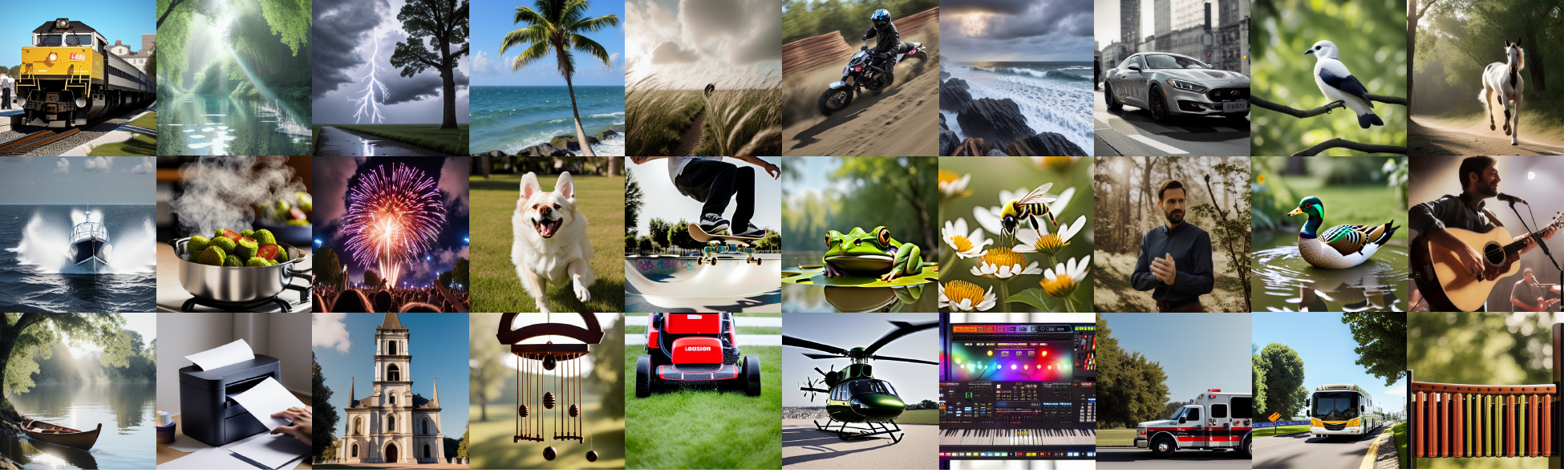}
    \caption{More results generated by AudioCanvas. Zoom in for a better viewing experience.}  
    \label{fig:good}  
\end{figure*}

In \cref{fig:comp}, we present visual comparisons between the baseline model and our AudioCanvas across both test datasets. The comparison shows that AudioCanvas outperforms other models through its strong visual effects, including the better use of lighting, shadow, and composition skill, and also high audio-visual alignment. In contrast, previous end-to-end trained models such as MACS struggle to produce satisfactory image generation, limited by the visual quality and alignment of their training data. Meanwhile, non-end-to-end models like AudioToken maintain good visual quality but frequently face semantic alignment issues. This leads them to sometimes generate images with meaningless content. Overall, our qualitative analysis results align closely with earlier objective metric evaluations, further validating several conjectures presented in the main article. In \cref{fig:good}, we present more pictures generated by AudioCanvas.

\section{Failure Cases}
\begin{figure*}[htbp]  
    \centering  
    \includegraphics[width=\textwidth]{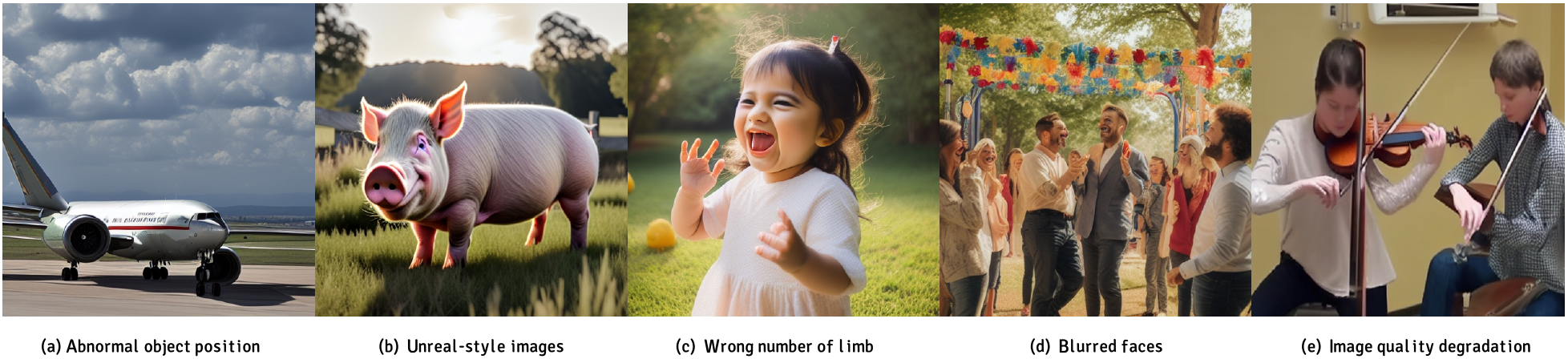}
    \caption{Some failure generated cases. Zoom in for a better viewing experience.}  
    \label{fig:bad}  
\end{figure*}

In this chapter, we further analyze some failed generation cases. In \cref{fig:bad}, we list five common cases of generation failure: (a) Abnormal object position: Images generated by the model may exhibit abnormalities in the positions of some objects. This may be because the backbone (SD 1.4) we use is relatively old. (b) Unreal-style images: AudioCanvas sometimes may generate Unreal-style images, which may be affected by the prior of images synthesized by FLUX.1-Krea-dev, although we have made every effort to avoid unrealistic-style images. (c) Wrong number of limbOverall, although limited by a relative weak backbone model, AudioCanvas still achieved much stronger generation performance than former A2I models. Although A2I is a complex and challenging task, we hope our research can better advance A2I research and ultimately achieve better generation results.

\end{document}